%% file: main.tex
\documentclass[10pt,twocolumn,letterpaper]{article}

\usepackage[pagenumbers]{cvpr}

\usepackage{times}
\usepackage{epsfig}
\usepackage{graphicx}
\usepackage{amsmath}
\usepackage{amssymb}
\usepackage{booktabs}
\usepackage[dvipsnames]{xcolor}
\usepackage{xcolor}
\usepackage{adjustbox}
\usepackage{multirow}
\usepackage{placeins}
\usepackage{enumitem}
\usepackage{subcaption}

\usepackage{color, colortbl}
\usepackage{listings}
\definecolor{codegreen}{rgb}{0,0.6,0}
\definecolor{codeblue}{rgb}{0,0,1}
\definecolor{codegray}{rgb}{0.5,0.5,0.5}
\definecolor{codepurple}{rgb}{0.58,0,0.82}
\definecolor{codered}{rgb}{0.73,0.08,0.08}
\definecolor{backcolour}{rgb}{1,1,1}
\definecolor{orange}{rgb}{0.98823529, 0.85098039, 0.7372549}
\newcommand{\cc}{\cellcolor{orange}}
\newcommand{\fdot}{{\Large $\bullet$}}

\usepackage{amssymb}
\usepackage{pifont}
\newcommand{\cmark}{\ding{51}}%
\newcommand{\xmark}{\ding{55}}%
\newcommand{\tinyp}[2][1]{\vspace{#1}\noindent\textbf{#2.}} 

\lstdefinestyle{mystyle}{
  backgroundcolor=\color{backcolour}, 
  commentstyle=\color{codered},
  keywordstyle=\color{codegreen}\textbf,
  numberstyle=\tiny\color{codegray}, 
  stringstyle=\color{codepurple},
  basicstyle=\ttfamily\small,
  breakatwhitespace=false,         
  breaklines=true,                 
  captionpos=b,                    
  keepspaces=true,                 
  numbers=none,                    
  numbersep=6pt,                  
  showspaces=false,                
  showstringspaces=false,
  showtabs=false,                  
  tabsize=2,
  basewidth={.5em},
}
\definecolor{cvprblue}{rgb}{0.21,0.49,0.74}
\usepackage[pagebackref,breaklinks,colorlinks,citecolor=cvprblue]{hyperref}

\usepackage[capitalize]{cleveref}
\crefname{section}{Sec.}{Secs.}
\Crefname{section}{Section}{Sections}
\Crefname{table}{Table}{Tables}
\crefname{table}{Tab.}{Tabs.}

\title{Early Action Recognition with Action Prototypes}

\author{
    Guglielmo Camporese$^{1}$\thanks{Work done while interning at AWS AI Labs.}\quad 
    Alessandro Bergamo$^{2}$\quad 
    Xunyu Lin$^{2}$ \quad 
    Joseph Tighe$^{2}$\thanks{Work done while at AWS AI Labs. At Meta now.}\quad 
    Davide Modolo$^{2}$\\
    $^{1}$University of Padova, Italy \quad $^{2}$AWS AI Labs\\
    {\tt\small $^{1}$guglielmo.camporese@phd.unipd.it\quad $^{2}$\{bergamo@, linxuny@, -@, dmodolo@\}amazon.com}
}

\begin{document}
\maketitle

\begin{abstract}
Early action recognition is an important and challenging problem that enables the recognition of an action from a partially observed video stream where the activity is potentially unfinished or even not started.
In this work, we propose a novel model that learns a prototypical representation of the full action for each class and uses it to regularize the architecture and the visual representations of the partial observations.
Our model is very simple in design and also efficient. We decompose the video into short clips, where a visual encoder extracts features from each clip independently. Later, a decoder aggregates together in an online fashion features from all the clips for the final class prediction.
During training, for each partial observation, the model is jointly trained to both predict the label as well as the action prototypical representation which acts as a regularizer.
We evaluate our method on multiple challenging real-world datasets and outperform the current state-of-the-art by a significant margin. For example, on early recognition observing only the first 10\% of each video, our method improves the SOTA by $+2.23$ Top-1 accuracy on Something-Something-v2, $+3.55$ on UCF-101, $+3.68$ on SSsub21, and $+5.03$ on EPIC-Kitchens-55, where prior work used either multi-modal inputs (e.g. optical-flow) or batched inference.
Finally, we also present exhaustive ablation studies to motivate the design choices we made, as well as gather insights regarding what our model is learning semantically.
\end{abstract}

\vspace{-6mm}
\section{Introduction}
\label{sec:intro}

\begin{figure}[t]
     \centering
     \includegraphics[width=\linewidth]{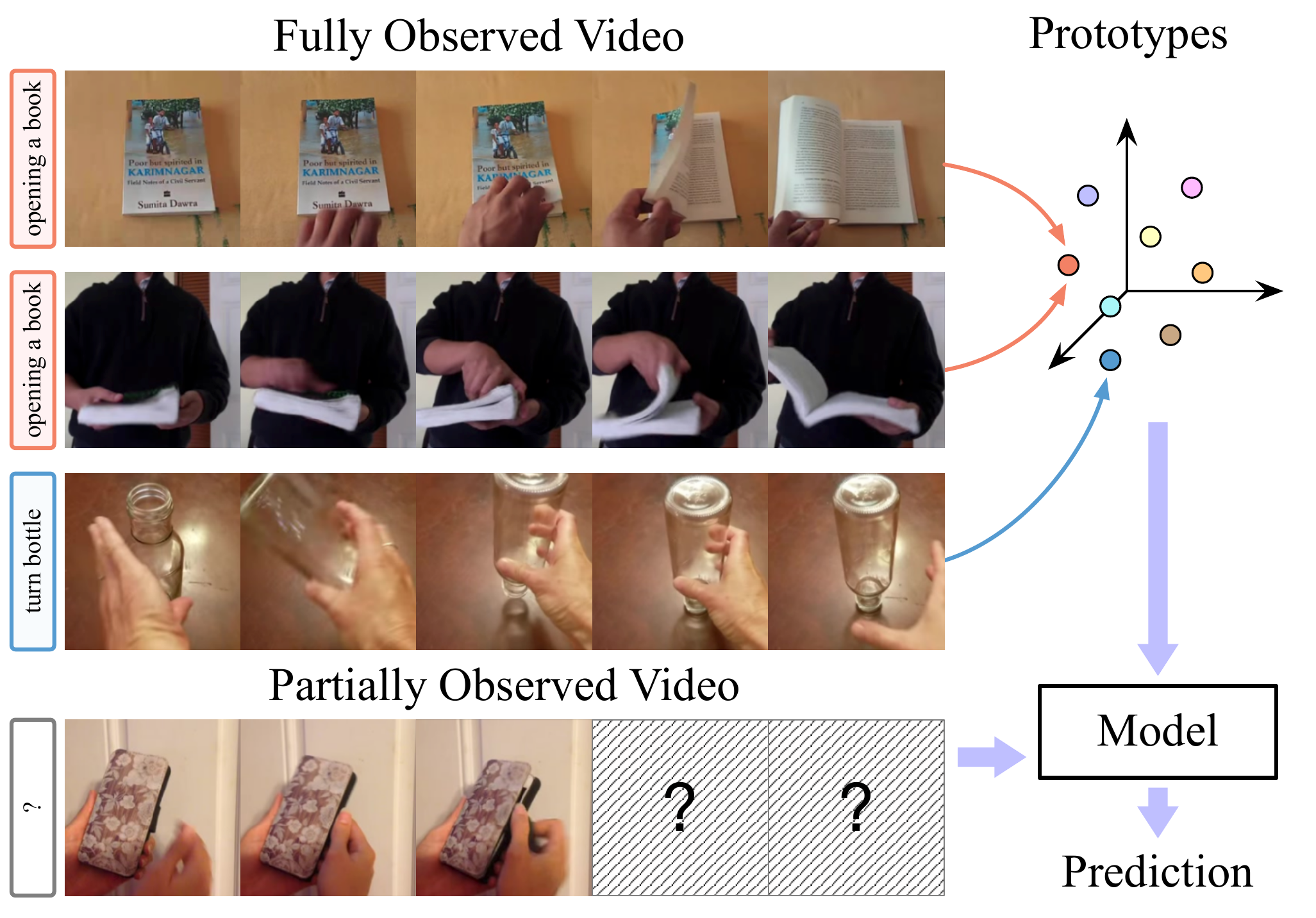}\vspace{-3mm}
     \caption{\small \it We propose to learn prototype presentations for each action class that summarize how the action typically evolves over time. During training, each prototype is initialized randomly and learned by observing the full action. The model is then simultaneously trained to recognize partially observed videos with the learned prototypes as reference.}
     \label{fig:idea}
     \vspace{-7mm}
\end{figure}

Our brain has an innate capability to anticipate events as they are about to happen. One of the fundamental ideas of the Predictive Coding (PC) theory~\cite{Millidge2021PredictiveCA} in neuroscience is that the brain is constantly engaged in predicting its upcoming states~\cite{Luczak2022, Ficco2021}. 
For example, if we are driving and see a pedestrian crossing our street, we immediately understand what would happen if we do not stop promptly~\cite{Rasouli2019PIEAL, Osman2022EarlyIP, Osman2022TAMFormerMT}. Moreover, when we see a chef in the kitchen grabbing a knife, we know they are likely going to cut something; and when we see a person approaching our door with a package in their hands, we know they are likely doing a package delivery. As humans, we are capable of making these preemptive predictions by only observing the initial phases of these situations. This ability is the key to enable several real-world applications, like pedestrian anticipation in autonomous driving, early recognition of activities from smart and multi-modal cameras, anticipation of possible dangerous situations from ego-centric cameras, and more. 

In this paper, we focus at helping action models to learn this ability automatically during training. This is the main goal of {\it early action recognition}, a topic that has recently started to get some important attention \cite{Wang2019ProgressiveTL, Zhao2019SpatiotemporalFR, Gammulle2019PredictingTF, Akbarian2017EncouragingLT}. Comparing to the classic action recognition approaches~\cite{Fan2021MultiscaleVT, Feichtenhofer2020X3DEA, Feichtenhofer2019SlowFastNF,Martnez2019ActionRW} which make predictions after observing the full actions and lack the ability to reason on partial observations (Fig.~\ref{fig:main_fig}, Table~\ref{tab:rec_ablation}), an early action recognition model is expected to predict the action contained in the entire video by looking only at the initial portion of it.
The models should thus have the ability to learn the action dynamics and the prior knowledge of how actions evolve over time. For example, if a video starts by showing a hand reaching for the kitchen utensils holder, the model needs to learn that ``opening a refrigerator'' is not a possible continuation, and the only actions this can evolve into are related to ``grabbing utensils''. As the model observes more of the video, it can then confidently predict the type of utensils grabbed (e.g., a spatula or a knife). 

Differently from previous works~\cite{Wang2019ProgressiveTL,Gammulle2019PredictingTF,Zhao2019SpatiotemporalFR, Furnari2021RollingUnrollingLF,Stergiou_2023_CVPR} that propose complex solutions that involve multiple training stages, multiple models or learning from multiple modalities, we propose a much simpler solution that models action dynamics as learnable prototypes. A prototype can be seen as a template of how an action typically evolves over time. As shown in Fig.~\ref{fig:idea}, a prototype of the action ``opening a book'' usually starts with a hand holding a book and slowly evolves into opening it. 
These prototypes are capable of representing the temporal dynamics of an action in a more holistic way compared to previous methods~\cite{Wang2019ProgressiveTL,Gammulle2019PredictingTF,Zhao2019SpatiotemporalFR} that often are too dependent on the visual content of a specific video. For example, our prototypes do not focus on learning what a book is, but rather about what a book is used for and how different dynamics lead to different potential outcomes (e.g., opening vs closing vs reading). We believe that this higher-level abstraction is the key to perform early action recognition. Thanks to this, our prototypes can suggest to reject unlikely actions and strengthen the confidence on those following a similar pattern. 

\begin{figure*}[ht]
     \centering
     \includegraphics[width=0.8\linewidth]{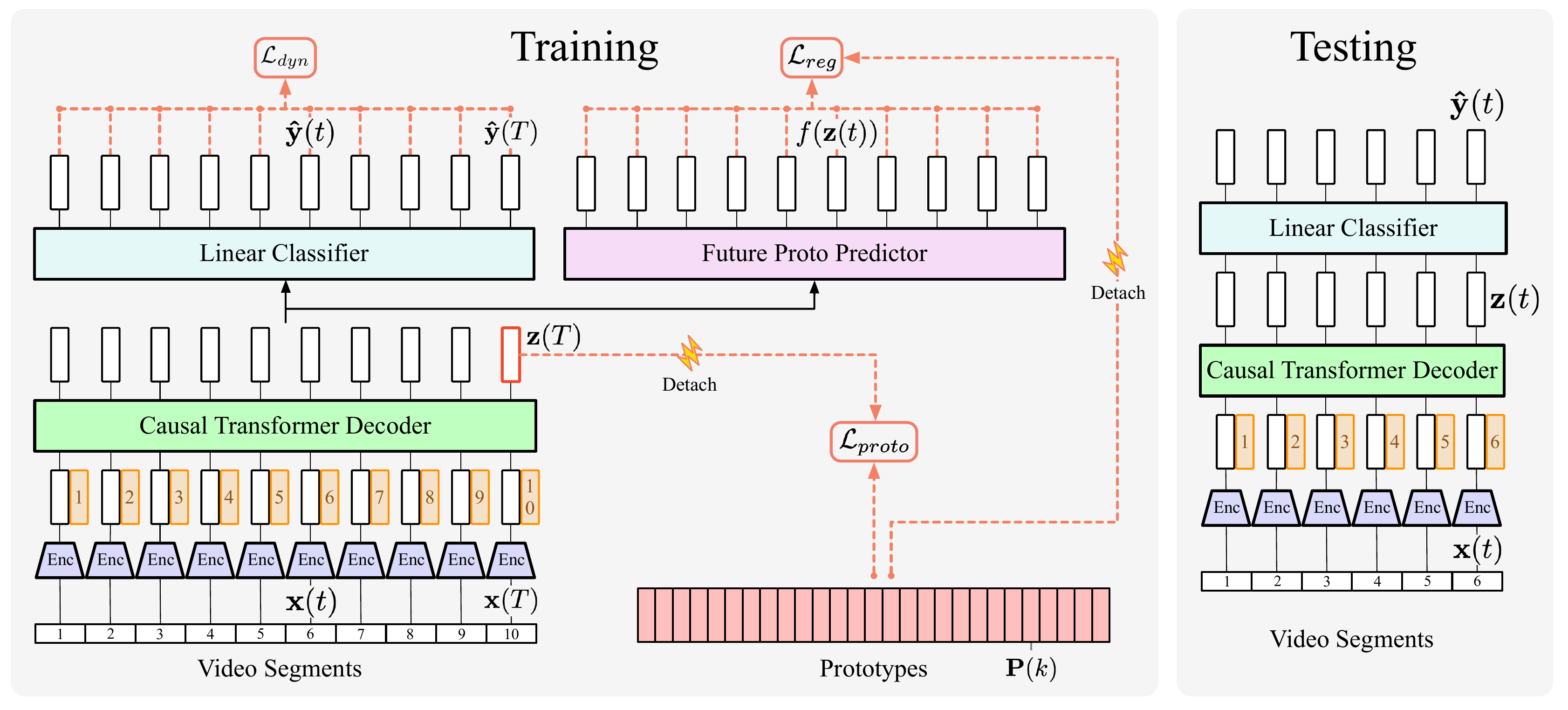}\vspace{-4mm}
     \caption{\small \it Overall architecture of our proposed model. The video is split into short segments that are independently encoded to feature vectors, to which positional embeddings are added. Later, a causal transformer decoder aggregates the input features and outputs a sequence of decoder features, which are subsequently fed to a classification head. 
     At training time (left figure), the model uses all the segments at once while using a causal attention mask to avoid peeking into the future.
     The architecture is regularized by our proposed prototypes, which are auxiliary learnable embeddings initialized from scratch. They model the prototypical representation of each action class. Each prototype can be seen as a cluster centroid, learned with a contrastive objective by minimizing ($\mathcal{L}_{proto}$ loss) the distance to the last decoder feature $\mathbf{z}(T)$ which captures the information of the whole video sequence. The regularization happens by constraining the decoder features for all the partial observations to not deviate from the prototypical representation of their corresponding ground-truth class ($\mathcal{L}_{reg}$ loss). The classification head is trained using our $\mathcal{L}_{dyn}$ loss. At test time, the segments are processed sequentially in an online fashion, and in the right figure we showcase the inference for the segment at $t=3$.
     }
     \label{fig:model}
     \vspace{-6mm}
\end{figure*}

To learn this prototypical representations, we propose Early-ViT (Fig.~\ref{fig:model}), which consists of an encoder-decoder architecture, followed by a prediction layer. During training, we empower Early-ViT with a memory bank of prototypes, which learn the prototypical representation of actions and enforces (through a regularization loss) the decoder to learn similar patterns. At inference, we then disregard the memory bank and directly predict actions using the decoder representation, which is now capable of reasoning about the temporal dynamic of each action and make predictions with limited observations.
Differently from the previous state-of-the-art TemPr~\cite{Stergiou_2023_CVPR}, that uses a batched inference mode and assumes that the number of segments is fixed and known a priori, our method  \emph{processes frames in an online fashion} and can therefore handle variable input segments: as a video segment is observed, the encoder feature vectors for the previous segments are re-utilized as-is and previous RGB frames are not needed. This is essential to enable action recognition in real-world live streaming applications and an important contribution of our method. 

We conduct experiments on the major action datasets: UCF-101~\cite{Soomro2012UCF101AD}, EPIC-Kitchens-55 ~\cite{Damen2021TheED} and Something-Something v2~\cite{Goyal2017TheS}. Our method consistently achieves state-of-the-art performance on all benchmarks, against previous works on early action recognition and classic action recognition (Fig.~\ref{fig:main_fig}).  Finally, we  present exhaustive ablation studies and qualitative visualizations to motivate our technical choices and gather insights on what the model learned.

\noindent In summary, we make the following contributions: 
\begin{itemize}[leftmargin=*,noitemsep,nolistsep] 
    \item We propose Early-ViT, a new architecture for early action recognition that uses a novel formulation to learn prototypical representations of actions and use them as a source of regularization.
    \item Early ViT is simple, yet effective. Compared to the previous SOTA ~\cite{Stergiou_2023_CVPR,Furnari2021RollingUnrollingLF}, our method trains one single unified model end-to-end, it uses single modality and its inference is considerably more efficient, as it uses a significant less amount of frames.
    \item We present SOTA results on several video datasets, under different settings. 
\end{itemize}

\vspace{-2mm}
\section{Related Work}
\label{sec:related_work}

\tinyp[0mm]{Video representation models}
Over the years, a lot of effort has been spent on modeling spatio-temporal representations of videos. Among these, we find hand-crafted video representations~\cite{Surasak2018HistogramOO, Dollr2005BehaviorRV, Efros2003RecognizingAA, Klser2008ASD, Laptev2008LearningRH, Peng2014ActionRW}, recurrent neural networks based architectures~\cite{Donahue2015LongtermRC, Jiang2019STMSA, Li2018RecurrentTP, Li2018VideoLSTMCA, Ng2015BeyondSS}, 2D CNN~\cite{Wang2015ActionRW, Wang2016TemporalSN, Wu2018CompressedVA}, and 3D CNN~\cite{Carreira2018ASN, Feichtenhofer2020X3DEA, Feichtenhofer2019SlowFastNF, Girdhar2019VideoAT, Li2018VideoLSTMCA, Qiu2017LearningSR}.

\tinyp[1mm]{Action recognition}
The general objective of the action recognition problem is to predict the action label of a given video that is fully-observed at once.
This is typically implemented by adding a classification layer on top of any video representation model~\cite{Fan2021MultiscaleVT, Feichtenhofer2020X3DEA, Feichtenhofer2019SlowFastNF,Martnez2019ActionRW}.
Note that this differs from our \textit{early} action recognition problem, where the video is only partially observed.

\tinyp[1mm]{Early action recognition and anticipation} 
In the early action recognition problem, the model has to recognize the action while observing only an initial portion of the video (see Fig.~\ref{fig:idea}).
Previous works on early action recognition investigated several approaches~\cite{Wang2019ProgressiveTL, Zhao2019SpatiotemporalFR, Gammulle2019PredictingTF, Akbarian2017EncouragingLT, Osman2022EarlyIP}, including ProgTS~\cite{Wang2019ProgressiveTL} that proposes a knowledge distillation framework where a teacher model, trained on fully observed videos, distills information to a student model, that processes only partially observed actions. Other works focused on regularizing the model by generating future representations in different ways: AA-GAN~\cite{Gammulle2019PredictingTF} employs a generative model, instead RGN-KF~\cite{Zhao2019SpatiotemporalFR} proposes to learn residual visual embeddings of future consecutive frames and to use a Kalman filter to correct future predictions. More recent approaches are RU-LSTM~\cite{Furnari2021RollingUnrollingLF} that proposes a multi-modal architecture that processes RGB frames as well as optical-flow and objects representations, and a TemPr~\cite{Stergiou_2023_CVPR} that early recognize the action in a multi-scale manner by processing the input video at different temporal resolutions. 

Similarly to early action recognition, in the action anticipation problem, the model can observe only a partial view of the video that precedes the starting point of the action. 
Previous works that tackle this problem are RU-LSTM~\cite{Furnari2021RollingUnrollingLF}, TAB~\cite{Sener2020TemporalAR}, AVT~\cite{Girdhar2021AnticipativeVT}, MeMViT~\cite{memvit}, and others~\cite{Osman2021SlowFastRL, Camporese2021KnowledgeDF}. In our work, we didn’t compare our proposed model to action anticipation models as we focused on the early action recognition problem that has different training and inference settings w.r.t. action anticipation.

\tinyp[1mm]{Prototype learning} 
Prototypical networks~\cite{Snell2017PrototypicalNF} learn to estimate the average representation of the labeled samples that belongs to different classes. The learned centroids are called prototypes, and their application is mostly on few-shot learning where the prototypes are used as references for predicting new inputs of unseen classes, and also for representing fine-grained spatio-temporal patterns~\cite{Martnez2019ActionRW}.

\vspace{-1.8mm}
\section{Method}
\label{sec:our_method}

We now present our method. First, we introduce our overall setting and model architecture in Sec.~\ref{sec:method_overall_architecture}.
Later, in Sec.~\ref{sec:method_use_action_prototypes} we describe our novel formulation of learning action prototypes which we use to regularize our decoder feature representations for all the partial observations.
Finally, in Sec.~\ref{sec:method_loss} we describe our novel dynamic loss for predicting accurately the action labels at all stages.

\vspace{-1mm}
\subsection{Overall architecture}
\label{sec:method_overall_architecture}
Figure~\ref{fig:model} depicts our model architecture. First, the video is split into short segments, which are fed to the model one at a time. The model uses an encoder to extract feature vectors from the individual video segments, and a temporal aggregator decoder that predicts the correct action label at each time. In the following paragraphs we describe in details these components.

\tinyp[1mm]{Model input}
The early recognition problem aims at recognizing actions starting from partially observed videos. Following previous works~\cite{Akbarian2017EncouragingLT, Gammulle2019PredictingTF, Zhao2019SpatiotemporalFR, Wang2019ProgressiveTL, Stergiou_2023_CVPR}, we define the partial observation ratio $\rho \in (0, 1]$ as the percentage of video that the model observes. More specifically, we partition a raw video of $F$ number of frames into $T$ non-overlapping clips, each having $F_c$ frames, obtaining the observation ratio defined as $\rho = t \cdot F_c/F$ with $t \in \{1, \dots, T\}$.

\vspace{-1mm}
\tinyp[1mm]{Visual encoder}
The visual backbone encoder processes one clip at a time. Specifically, given the $t$-th video clip $\mathbf{x}(t) \in \mathbb{R}^{C \times F_c \times H \times W}$ with $F_c$ number of frames, the visual encoder produces a $D_{enc}$ dimensional feature representation that is subsequently projected with a linear layer into $\mathbf{z_{enc}}(t) \in \mathbb{R}^{D}$ in order to be compatible with the decoder. Our encoder design is general and does not require a specific architecture, however, in our investigation we adopted the recent MViT~\cite{Fan2021MultiscaleVT} since it is an efficient and strong transformer based model.

\vspace{-1mm}
\tinyp[1mm]{Temporal aggregator decoder}
The decoder takes as input the output sequence of the encoder $\mathbf{z_{enc}} \in \mathbb{R}^{T \times D}$, it adds the learnable positional embeddigs, and it produces the decoder features $\mathbf{z} \in \mathbb{R}^{T \times D}$.
In our investigation we use a Transformer Decoder~\cite{NIPS2017_3f5ee243}, and to avoid the processing of future tokens during the computation, the decoder adopts a masked multi-head self-attention with causal masks. Subsequently, the final action prediction $\hat{y}$ is computed by a linear head $h(\cdot)$ that processes each token representation of the decoder independently as $\hat{y}(t) = h(\mathbf{z}(t))$, optimizing the classification objective:

\vspace{-2mm}
\begin{equation}
    \mathcal{L}_{clf}(\mathbf{x}, y, t) = - \sum_{k=1}^{K} y(k) \log\hat{y}(t, k),
\end{equation}

where $K$ is the number of action classes of the dataset.

\subsection{Use action prototypes as model regularization}
\label{sec:method_use_action_prototypes}

\tinyp[0mm]{Learning prototypes}
The prototypes are learnable embeddings $\mathbf{P} \in \mathbb{R}^{K \times D}$ that encode the prototypical representation of each action. In order to train the prototypes, we use the decoder feature vector of the last segment $\mathbf{z}(T) \in \mathbb{R}^D$ as it encodes the whole spatio-temporal discriminative information of the action.
Specifically, the representation $\mathbf{z}(T)$ is compared with all the $K$ prototypes under the $\ell_2$ similarity as:

\vspace{-5mm}
\begin{equation}
    s(k) = - \left\| \mathbf{P}(k) - sg\big( \mathbf{z}(T) \big) \right\|,
\end{equation}

where $s(k) \in \mathbb{R}$ and $\mathbf{P}(k) \in \mathbb{R}^D$ are respectively the similarity score and the prototype of the $k$-th class, and $sg(\cdot)$ is the stop-gradient operator. The similarity scores are softmax-activated and converted to a class distribution that can be trained with the cross-entropy loss. In particular, we define the objective for learning the prototypes as:

\vspace{-5mm}
\begin{equation}
    \mathcal{L}_{proto}(\mathbf{x}, y) = - \sum_{k=1}^K y(k) \log\left( \frac{e^{s(k)}}{\sum_{j=1}^K{e^{s(j)}}} \right)
\end{equation}

where the similarity distribution is encouraged to be similar to the one-hot representation of the action class. Specifically, with this loss we maximize the similarity between the decoder representation $\mathbf{z}(T)$ and the prototype of the ground truth label $\mathbf{P}(y)$, and at the same time we minimize the similarity between the feature representation $\mathbf{z}(T)$ and all the other prototypes. This constrastive objective is implicitly defined in the cross-entropy loss. In order to learn the prototypes, we detach the target signal $\mathbf{z}(T)$ on the computation of $\mathcal{L}_{proto}$ with the stop-gradient operation for separating the prototype objective from the other ones and not experiencing collapses issues.

It's worth noting that we differentiate our prototypes design from~\cite{Snell2017PrototypicalNF} as we store the prototypes in a separate memory bank that can be updated and accessed during training and inference whereas in the original paper the prototypes need to be estimated at every iteration during training by averaging embeddings of the same class, leading to a slower training compared to our solution. Moreover, we simplify the training as we employ a standard random sampling of the input, whereas in~\cite{Snell2017PrototypicalNF} the authors need to construct batches by sampling inputs in the same class in order to have a solid estimate of the centroids. However, this sampling procedure could introduce a training bias when also other objectives are optimized together with the prototypes learning.

\tinyp[2mm]{Regularization predicting future prototypes}
One way to leverage the prototypes during the model training is to regularize the architecture by constraining the output feature space of the decoder to the regions where it is possible to predict the action prototypes. More specifically, we defined a feature predictive module $f(\cdot)$ that tries to predict the correct prototype from decoders' embeddings $\mathbf{z}(t)$ at all the partial observation steps $t$. In particular, we compute the $\ell_2$ similarity $s_{reg}$ between the predicted feature $f(\mathbf{z}(t))$ and all the prototypes $\mathbf{P}$ as:

\vspace{-3mm}
\begin{equation}
    s_{reg}(t, k) = - \left\| sg\big( \mathbf{P}(k) \big) - f(\mathbf{z}(t)) \right\|
\end{equation}

and we optimize the following objective:

\vspace{-5mm}
\begin{equation}
    \mathcal{L}_{reg}(\mathbf{x}, y, t) = - \sum_{k=1}^K y(k) \log \left( \frac{e^{s_{reg}(t, k)}}{\sum_{j=1}^K{e^{s_{reg}(t, j)}}} \right)
\end{equation}

The feature predictive module is implemented as a multi-layer perceptron that processes the decoder feature vectors at each partial observation independently.
In order to train the feature predictive module we detach the prototype target signal $\mathbf{P}$ with the stop-gradient operator for separating the regularization objective from the other ones and not experiencing collapses issues as done in the prototypes objective.

\subsection{Temporal Losses}
\label{sec:method_loss}
In this section we introduce a novel dynamic loss that helps the model learning to predict the correct action both when the video is partially observed as well as fully observed.
We found that when the model is trained with a pure action classification loss, it performs accurately on action recognition of fully observed videos, but not so well on partial observations (i.e., at the early temporal stages).
In contrast, training the model with a loss that optimizes  all the temporal steps leads to better accuracy at the early stages but degraded accuracy in the final stages.
To balance these and achieve the best performance throughout, we develop a new dynamic loss that combines the benefits from both temporal losses. Formally, we define the action recognition loss that optimizes \emph{only the last} temporal step as:

\vspace{-3mm}
\begin{equation}
    \mathcal{L}_{ol}(\mathbf{x}, y) = \mathcal{L}_{clf}(\mathbf{x}, y, T)
    \label{eq:L_ol}
\end{equation}
and the loss that trains \emph{all} the temporal tokens:
\vspace{-3mm}
 \begin{equation}
    \mathcal{L}_{all}(\mathbf{x}, y) = \frac{1}{T}\sum_{t=1}^T \mathcal{L}_{clf}(\mathbf{x}, y, t).
    \label{eq:L_all}
\end{equation}

To balance the trade-off of the two previously mentioned temporal objectives, we propose our dynamic loss that benefits from $\mathcal{L}_{ol}$'s high final accuracy early in the training process to shape the embedding and prototype training, but it also benefits from $\mathcal{L}_{all}$ later to achieve high accuracy in early predictions.
More specifically, our loss is defined as:

\begin{equation}
  \mathcal{L}_{dyn}(\mathbf{x}, y) =
    \begin{cases}
      \mathcal{L}_{ol}(\mathbf{x}, y) & \text{if $e \leq e^*$}\\
      \mathcal{L}_{all}(\mathbf{x}, y) & \text{if $e > e^*$}\\
    \end{cases}  
    \label{eq:L_dyn}
\end{equation}

where $e$ is the training epoch and $e^*$ is the switch epoch.

The final total loss is the sum of the previously mentioned losses that are related to the early action recognition ($\mathcal{L}_{dyn}$), the learning of the prototypes ($\mathcal{L}_{proto}$), and the training of the future predictor ($\mathcal{L}_{reg}$):

\vspace{-3mm}
\begin{equation}
    \mathcal{L}_{tot} = \mathcal{L}_{dyn} + \mathcal{L}_{proto} + \mathcal{L}_{reg}
    \label{eq:L_tot}
\end{equation}

\vspace{-4mm}
\section{Experiments}
\label{sec:experiments}

\vspace{-2mm}
\tinyp[1mm]{Datasets}
We conduct experiments on several datasets: UCF-101~\cite{Soomro2012UCF101AD}, EPIC-Kitchens-55 (EK55)~\cite{Damen2021TheED}, Something-Something v2~\cite{Goyal2017TheS} (SSv2) and a reduced version of Something Something (SSsub21) used in previous works~\cite{Akbarian2017EncouragingLT, Kong2018ActionPF, Wu2021AnticipatingFR, Wu2021SpatialTemporalRR, Stergiou_2023_CVPR}. UCF-101~\cite{Soomro2012UCF101AD} consists of $13$k videos in total of $101$ action classes. The dataset includes various types of human actions including human-object interactions, body-motions, human-human interactions, playing musical instruments and sports. As a common practice, we report the results on the first split of the validation dataset. EPIC-Kitchens-55~\cite{Damen2021TheED} is an egocentric video dataset that contains $40$k action segments of $2513$ classes. Each action class is composed by a single verb and noun pair coming from a set of $125$ unique verbs, and $352$ unique nouns. We use the train and validation dataset split provided by~\cite{Furnari2021RollingUnrollingLF}. The Something-Something v2 dataset~\cite{Goyal2017TheS} contains $169$k training, and $25$k validation videos. The videos show human-object interactions to be classified into $174$ classes which is known as a ‘temporal modeling‘ task, and we report the accuracy on the validation set. Moreover, we use also a subset of the Something Something dataset containing $13$k video samples of $21$ classes for comparing with previous works.

\tinyp[1mm]{Implementation details}
For all the experiments with our model, we split the videos into 10 non-overlapping clips each having 2 frames (with the exception of UCF-101 for which we use 4 frames). Frames are randomly rescaled, resulting in their shorted side to be between 256 and 320 pixels, and finally cropped to 224 pixels. As further data augmentation, we use 2 layers of RandAugment~\cite{RandAugment} with 0.5 of probability.
We use MViT-B-16x4~\cite{Fan2021MultiscaleVT} as visual encoder,  pre-trained on Kinetics-400~\cite{Kay2017TheKH}.
The decoder, named "T-Dec-B" in this work, is implemented with a sequence of six Transformer encoder blocks as present in~\cite{NIPS2017_3f5ee243} with the only modification of adding a causal mask in the multi-head attention module.
We optimize the model using AdamW~\cite{adamw} for 30 epochs, and set $e^*=15$ in $\mathcal{L}_{dyn}$. We do apply label smoothing. We have one prototype per each action class, implemented as 256-dimensional embeddings vectors which are randomly initialized before training.
The training step implementation, including stop-gradient operations, are presented in PyTorch-like code in the supplementary material.

\tinyp[1mm]{Evaluation metrics} To measure the performance of our early action recognition model we follow previous works; at each partial observation of the input video the model is evaluated with the standard classification accuracy. Moreover, not only we are interested in maximizing the overall accuracy of the model, but we also focused on recognizing the action as early as possible.
Following previous works, we measure the early recognition capability of the model with the Area Under the Curve (AUC) defined by the partial observation ratios and the accuracy curves. This metric aggregates all the accuracy information of different steps in a single scalar value.

\begin{table*}[!htp]
    \centering
    \begin{adjustbox}{width=1.0\linewidth,center}
	\setlength{\tabcolsep}{5pt}
    \begin{tabular}{l c c c c c c c c c c c c c c c c} \toprule
    \multirow{2}{*}{\textbf{Model}} & \multirow{2}{*}{\textbf{Conference}} & \multirow{2}{*}{\textbf{Encoder}} & \multirow{2}{*}{\textbf{Decoder}} & \multirow{2}{*}{\textbf{\shortstack[c]{Inference\\ Mode}}} &  \multicolumn{10}{c}{\textbf{Top-1 Acc}}  & \multirow{2}{*}{\textbf{AUC}}\\
    & & & & & \textbf{10\%} & \textbf{20\%} & \textbf{30\%} & \textbf{40\%} & \textbf{50\%} & \textbf{60\%} & \textbf{70\%} & \textbf{80\%} & \textbf{90\%} & \textbf{100\%} & \\
    \midrule
    TemPr~\cite{Stergiou_2023_CVPR} & CVPR-2023 & V-Swin-B & TemPr & Batched & 20.5 &	-	& 28.6	& -	& 41.2	& -	& 47.1	& - &	-	& -	& - \\
    Early-ViT (ours) &  - & V-Swin-B & T-Dec-B & Online & \cc \textbf{23.34}	& 27.0	& 31.93	& 37.84	& 44.47	& 50.67	& 55.35	& 58.81	& 60.75	& 61.33	& 40.92 \\
    Early-ViT (ours)&  - & MViT-B &	T-Dec-B & Online & 22.73	& \cc \textbf{27.81}	& \cc \textbf{33.62}	& \cc \textbf{40.52}	& \cc \textbf{47.95}	& \cc \textbf{53.94}	& \cc \textbf{58.54}	& \cc \textbf{61.49}	& \cc \textbf{63.03}	& \cc \textbf{63.56}	& \cc \textbf{43.00}	 \\
    \bottomrule
    \end{tabular}
    \end{adjustbox}\vspace{-3mm}
    \caption{\small \it Top-1 accuracy for different video observation ratios on the Something Something v2 dataset, which contains 174 classes.}
    \label{tab:ssv2_results}
\end{table*}

\begin{table*}[!htp]
    \centering
    \begin{adjustbox}{width=1.0\linewidth,center}
	\setlength{\tabcolsep}{3pt}
    \begin{tabular}{l c c c c c c c c c c c c c c} \toprule
    \multirow{2}{*}{\textbf{Model}} & \multirow{2}{*}{\textbf{Conference}} & \multirow{2}{*}{\textbf{Encoder}} & \multirow{2}{*}{\textbf{Decoder}} &  \multirow{2}{*}{\textbf{\shortstack[c]{Inference\\ Mode}}} & \multicolumn{8}{c}{\textbf{Top-1 Acc}}  & \multirow{2}{*}{\textbf{AUC}}\\
    & & & & & \textbf{12.5\%} & \textbf{25\%} & \textbf{37.5\%} & \textbf{50\%} & \textbf{62.5\%} & \textbf{75\%} & \textbf{87.5\%} & \textbf{100\%} & \\
    \midrule
    RU-LSTM~\cite{Furnari2021RollingUnrollingLF} (rgb) & ICCV-2021 & TSN & RU-LSTM & Online & 20.09 &	22.97	& 24.38	& 25.54	& 26.75	& 27.49	& 27.84 & 28.32 &	22.40 \\
    RU-LSTM~\cite{Furnari2021RollingUnrollingLF} (rgb + obj + flow) & ICCV-2021 & TSN & RU-LSTM & Online & 24.48 & 27.63	& 29.44	& 30.93	& 32.16	& 33.09	& 33.63 & 34.07 & 27.02 \\
    RU-LSTM$^*$~\cite{Furnari2021RollingUnrollingLF} (rgb) & ICCV-2021 & 
 MViT-B & RU-LSTM & Online & 21.23 & 23.95 & 25.65 & 26.75 & 27.92 & 28.6 & 28.92 & 29.54 & 23.40 \\
    RU-LSTM$^*$~\cite{Furnari2021RollingUnrollingLF} (rgb + obj + flow) & ICCV-2021 & MViT-B & RU-LSTM & Online & 22.08 & 24.65 & 27.09 & 28.52 & 28.76 & 29.24 & 29.88 & 30.52 & 24.30 \\
    Early-ViT (ours)&	- & MViT-B &	T-Dec-B &  Online & \cc \textbf{25.12}	& \cc \textbf{28.11}	& \cc \textbf{30.36}	& \cc \textbf{32.1}	& \cc \textbf{33.93}	& \cc \textbf{35.25}	& \cc \textbf{35.71}	& \cc \textbf{36.34}	& \cc \textbf{28.27} \\
    \bottomrule
    \end{tabular}
    \end{adjustbox}\vspace{-3mm}
    \caption{\small \it Top-1 acc. for different video observations on the challenging EPIC-Kitchens-55, which has 2,513 classes. * indicates our run.}
    \label{tab:ek55_results}
    \vspace{-2mm}
\end{table*}

\begin{table*}[!htp]
    \centering
    \begin{adjustbox}{width=1.0\linewidth,center}
	\setlength{\tabcolsep}{4pt}
    \begin{tabular}{l c c c c c c c c c c c c c c c} \toprule
    \multirow{2}{*}{\textbf{Model}} & \multirow{2}{*}{\textbf{Conference}} & \multirow{2}{*}{\textbf{Encoder}} & \multirow{2}{*}{\textbf{Decoder}} & \multirow{2}{*}{\textbf{\shortstack[c]{Inference \\ Mode}}} & \multicolumn{10}{c}{\textbf{Top-1 Acc}}  & \multirow{2}{*}{\textbf{AUC}}\\
    & & & & & \textbf{10\%} & \textbf{20\%} & \textbf{30\%} & \textbf{40\%} & \textbf{50\%} & \textbf{60\%} & \textbf{70\%} & \textbf{80\%} & \textbf{90\%} & \textbf{100\%} & \\
    \midrule
    ProgTS~\cite{Wang2019ProgressiveTL} & CVPR-2019 & ResNext-101 & LSTM & Online & 83.32 &	87.13 &	88.92 &	89.82 &	90.85 &	91.04 &	91.28 &	91.23 &	91.31 &	91.47 &	80.9 \\
    RGN-KF~\cite{Zhao2019SpatiotemporalFR} & ICCV-2019 & TSN & RGN & Online &  83.1 & 85.16 &	- &	90.78 &	- &	92.03 &	- &	93.19 &	- &	- &	- \\
    AA-GAN~\cite{Gammulle2019PredictingTF} & ICCV-2019 & ResNet-50 & LSTM & Online & - &	84.2 &	- &	- &	85.6 &	- &	- &	- &	- &	- &	- \\

    TemPr~\cite{Stergiou_2023_CVPR} &	CVPR-2023 & ResNet-50 (3D) & TemPr & Batched & 84.8 & 90.5 &	91.2 &	91.8 &	91.9 &	92.2 &	92.3 &	92.4 &	92.6 &	- &	-\\
    TemPr$^\dag$~\cite{Stergiou_2023_CVPR} &	CVPR-2023 & MoViNet-A4 &	TemPr &	Batched & 88.6 &	93.5 &	94.9 &	94.9 &	95.4 &	95.2 &	95.3 &	\cc \textbf{96.6} &	96.2 &	- &	- \\
    TemPr$^*$~\cite{Stergiou_2023_CVPR} &	CVPR-2023 & MViT-B &	TemPr & Batched & 85.7 &	- &	87.4 &	- &	87.5 &	- &	86.9 &	- &	86.5 &	- &	-\\
    Early-ViT (ours) & - & ResNet-50 (3D) & T-Dec-B & Online & 87.5 & 90.5 & 91.9 & 92.2 & 92.8 & 93.2 & 93.4 & 94.0 & 94.0 & 94.0 & 83.3 \\
    Early-ViT (ours) & - & MoViNet-A4 &	T-Dec-B & Online & 87.16 & 90.14 & 91.68 & 92.23 & 92.86 & 93.37 & 93.58 & 93.47 & 93.5 & 93.39 & 83.11 \\
    Early-ViT (ours) &	- & MViT-B & T-Dec-B & Online & \cc \textbf{92.15}	& \cc \textbf{93.97}	& \cc \textbf{95.03}	& \cc \textbf{95.3}	& \cc \textbf{95.75}	& \cc \textbf{95.88}	& \cc \textbf{96.35} &  96.33	& \cc \textbf{96.54}	& \cc \textbf{96.64}	& \cc \textbf{85.95} \\
    \bottomrule
    \end{tabular}
    \end{adjustbox}
    \vspace{-3mm}
    \caption{\small \it Experiments on UCF-101. Our method achieves SOTA results, and improves over the prior work~\cite{Stergiou_2023_CVPR} that uses batched inference. Using the original code of~\cite{Stergiou_2023_CVPR} we were not able to reproduce the results of $\dag$ and * was obtained by building on the original code of~\cite{Stergiou_2023_CVPR}.}
    \label{tab:ucf101_results}
    \vspace{-2mm}
\end{table*}

\begin{table*}[!htp]
    \centering
    \begin{adjustbox}{width=1.0\linewidth,center}
	\setlength{\tabcolsep}{4pt}
    \begin{tabular}{l c c c c c c c c c c c c c c c} \toprule
    \multirow{2}{*}{\textbf{Model}} & \multirow{2}{*}{\textbf{Conference}} & \multirow{2}{*}{\textbf{Encoder}} & \multirow{2}{*}{\textbf{Decoder}} &  \multirow{2}{*}{\textbf{\shortstack[c]{Inference\\ Mode}}} & \multicolumn{10}{c}{\textbf{Top-1 Acc}}  & \multirow{2}{*}{\textbf{AUC}}\\
    & & & & & \textbf{10\%} & \textbf{20\%} & \textbf{30\%} & \textbf{40\%} & \textbf{50\%} & \textbf{60\%} & \textbf{70\%} & \textbf{80\%} & \textbf{90\%} & \textbf{100\%} & \\
    \midrule
    MS-LSTM~\cite{Akbarian2017EncouragingLT} & ICCV-2017 & VGG-16 & MS-LSTM & Online & 16.9 & 16.6 & 16.8 & - & 16.7 & - & 16.9 & - & 17.1 & - & -\\
    mem-LSTM~\cite{Kong2018ActionPF} & AAAI-2018 & ResNet-18 & mem-LSTM & Online & 14.9 & 17.2 & 18.1 & - & 20.4 & - & 23.2 & - & 24.5 & - & -\\
    MSRNN~\cite{Akbarian2017EncouragingLT} & ICCV-2017 & VGG-16 & S-RNN & Online & 20.1 & 20.5 & 21.1 & - & 22.5 & - & 24.0 & - & 27.1 & - & -\\
    GGN+LSTGCNN~\cite{Wu2021AnticipatingFR} & AAAI-2021 & ResNet-50 & GNN & Online & 21.2 & 21.5 & 23.3 & - & 27.4 & - & 30.2 & - & 30.5 & - & -\\
    IGGN+LSTGCNN~\cite{Wu2021SpatialTemporalRR} & IJCV-2021 & ResNet-50 & GCN & Online & 22.6 & - & 25.0 & - & 28.3 & - & 32.2 & - & 34.1 & - & -\\
    TemPr$^\dag$~\cite{Stergiou_2023_CVPR} & CVPR-2023 & MoViNet-A4 & TemPr & Batched & 28.4 & 34.8 & 37.9 & - & 41.3 & - & 45.8 & - & 48.6 & - & -\\
    Early-ViT$^*$ (ours) & - & MoViNet-A4 & T-Dec-B & Online & 23.9 & 28.8 & 33.3 & 38.9 & 45.0 & 48.1 & 51.9 & 54.0 & 54.5 & 54.7 & 39.38	 \\
    Early-ViT (ours) & - & MViT-B &	 T-Dec-B & Online & \cc \textbf{32.08}	& \cc \textbf{35.52}	& \cc \textbf{40.31}	& \cc \textbf{47.39}	& \cc \textbf{52.36}	& \cc \textbf{57.14}	& \cc \textbf{60.65}	& \cc \textbf{61.86}	& \cc \textbf{62.24}	& \cc \textbf{62.12}	& \cc \textbf{46.46}	 \\
    \bottomrule
    \end{tabular}
    \end{adjustbox}\vspace{-3mm}
    \caption{\small \it Top-1 acc. for different video observation ratios on the Something Something sub21, which contains 21 classes. $*:$ For apple-to-apple comparison, we trained Early-ViT with MoViNet-A4. 
    Further comparison with TemPr including a plot on the accuracy vs the amount of frames used is provided in the supplementary material. Using the original code of~\cite{Stergiou_2023_CVPR} we were not able to reproduce the results of $\dag.$}
    \label{tab:sssub21_results}
    \vspace{-4mm}
\end{table*}

\vspace{-2mm}
\subsection{Comparison against the state-of-the-art}

\vspace{-1mm}
\tinyp[0mm]{``Online" vs ``Batched" inference}
In our comparison against previous works on early action recognition, we divide the modeling frameworks into two different categories in relation to their inference and training processing: \textit{online} vs \textit{batched}. Specifically, The \textit{online} inference mode has one model to produce \textit{all} predictions for a varied amount of partial observations from an online video stream (as described in Sec.~\ref{sec:method_overall_architecture}).
Instead, \textit{batched} inference mode assumes a fixed number of partial observations known a priori, as it trains a separate model \textit{for each} observation ratio, resulting in having a total of $T$ models. This means that the method cannot handle a variable number of segments at test time, and it also requires that the previous frames must be kept in memory which is not desirable for streaming applications.
Our method adopts the \textit{online} mode by training only one model end-to-end and testing on a variable number of video segments in an online fashion. It greatly simplifies the training and deployment complexity over the batched mode.

\tinyp[0mm]{Against early-action recognition models}
As reported in Table~\ref{tab:ssv2_results} our model achieves better results on the large-scale SSv2 dataset w.r.t. to TemPr~\cite{Stergiou_2023_CVPR} on all partial observations using the same backbone (V-Swin-B), gaining $2.84$ points of accuracy at the early observation ratio $\rho = 10\%$. Furthermore, we can boost the performance by using the M-ViT, which interestingly is a weaker visual backbone than V-Swin-B, gaining $5.02$ points at $\rho = 50\%$ w.r.t.~\cite{Stergiou_2023_CVPR}.
It is worth noting that TemPr~\cite{Stergiou_2023_CVPR} adopts the \textit{batched} inference mode. As segment $t$ is observed, TemPr samples a new batch of frames from all the segments observed until $t$, and performs inference using the $t$-th model. 
In contrast, our method uses a \emph{single model and processes frames in an online fashion}. As a video segment is observed, the encoder feature vectors for the previous segments are re-utilized as-is, hence previous RGB frames are not needed. In addition, although we tested our model on a fixed number of segments for comparison with the prior work, our decoder naturally handles a variable number of segments. Furthermore, the multi-scale design in TemPr~\cite{Stergiou_2023_CVPR} samples a total of $T*S*F$ frames, where $S=4$ is the number of scales and $F=16$ is the number of frames to sample for each scale, which is significantly more than our method (single scale $S=1$, $F\in\{2,4\}$).
On the EK-55 dataset we compare with RU-LSTM~\cite{Furnari2021RollingUnrollingLF}, a multi-modal encoder-decoder architecture (Table~\ref{tab:ek55_results}). Our method improves the results by $5.03$ Top-1 accuracy points at $\rho=12.5\%$ and by $5.87$ points AUC when using the RGB input. Interestingly, our method that processes a single modality (RGB) is also more accurate than the multi-modal RU-LSTM that processes RGB, Optical Flow, and pre-computed object predictions. For a fair comparison, we re-trained RU-LSTM with the same encoder (MViT-B), and our method still outperforms it by $3.97\%$ AUC and around $3\%$ to $6\%$ of absolute Top-1 accuracy improvements across all observation ratios.
In Table~\ref{tab:ucf101_results} we reported the results on UCF-101. We compared our method with previous works, and to make a fair comparison with the SOTA~\cite{Stergiou_2023_CVPR} we trained our model employing the same visual backbones. We showed that using the 3D ResNet-50 and the MViT-B encoders our model is more accurate on \textit{all} the observation ratios. However, our Early-ViT equipped with the MoViNet-A4 backbone does not outperform TemPr with the same encoder and this is due to the fact that the MoViNet-A4 is optimized to work
on large inputs (i.e., TemPr uses $4 \times 16 = 64$ frames for each partial observation) and is not adequate as-is for shorter ones (Early-ViT uses $2$ frames for each observation).
Moreover, we found reproducibility issues in the original codebase from TemPr~\cite{Stergiou_2023_CVPR} as we were not able to achieve the originally reported accuracy of the model equipped with MoViNet-A4 on UCF-101 (Table~\ref{tab:ucf101_results}) and on SSsub21 (Table~\ref{tab:sssub21_results}).
Finally, we reported the results on the SSsub21 dataset in Table~\ref{tab:sssub21_results}. We gain $3.68$ points at the early stage $\rho=10\%$ with respect to TemPr~\cite{Stergiou_2023_CVPR}, and $13.64$ at the final stages $\rho=90\%$.

\begin{figure}[t]
     \centering
     \includegraphics[width=0.7\linewidth]{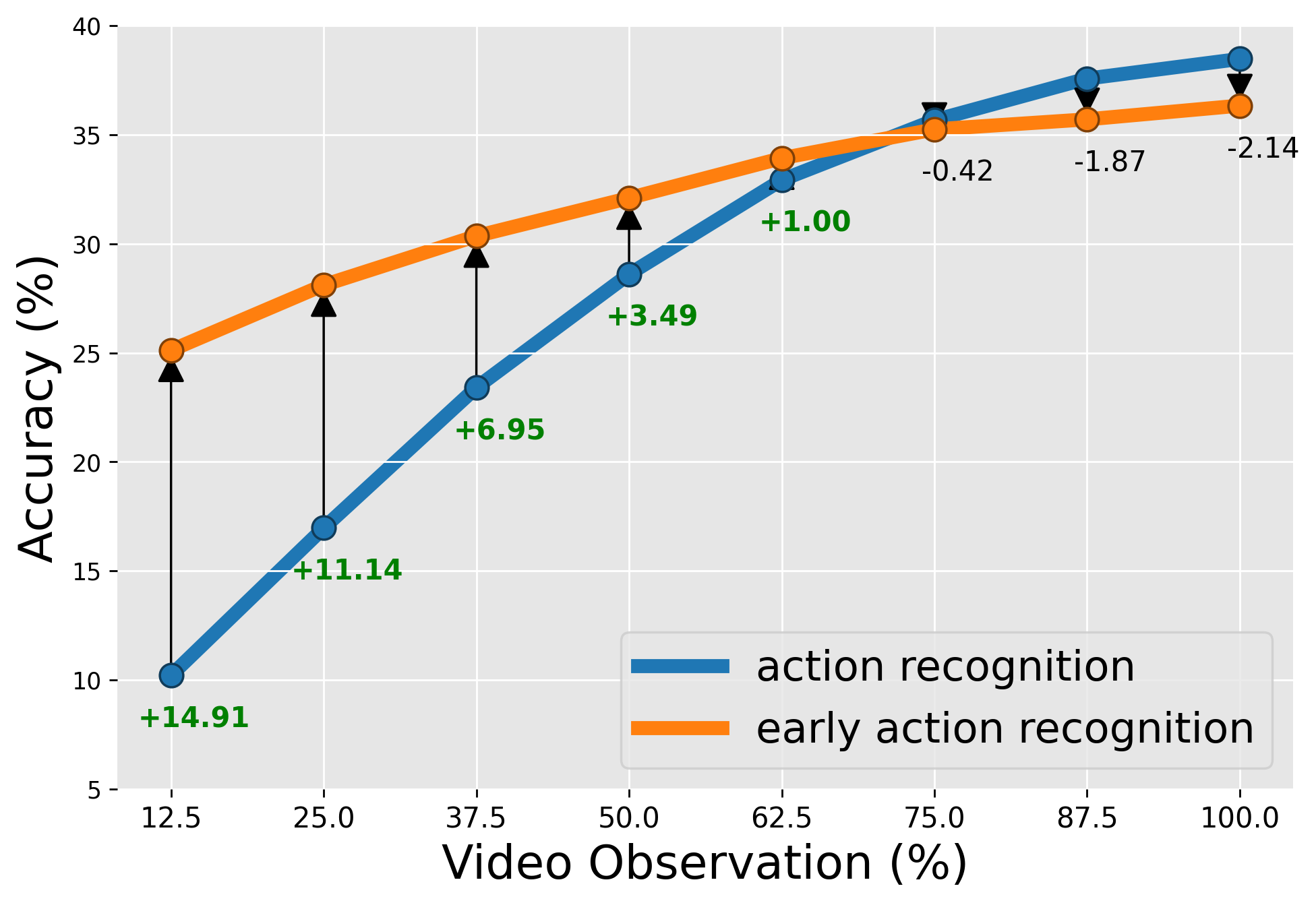} \vspace{-3mm}
     \caption{\small \it Validation accuracy on the challenging EPIC-Kitchens-55~\cite{Damen2021TheED} dataset as a function of the ratio of video that was observed. Standard action recognition models (e.g. MViT~\cite{Fan2021MultiscaleVT} shown in blue color) lack in accuracy when the action is partially observed, whereas our model (shown in orange color) designed for early action recognition highly improves the results at early times.}
     \label{fig:main_fig}
     \vspace{-7mm}
\end{figure}

\tinyp[2mm]{Against classic full-video recognition models}
To demonstrate the effectiveness of our method on early action recognition and the fact that high-accuracy cannot be simply achieved by utilizing the standard action recognition training scheme, we compare the following three methods on EPIC-Kitchens-55:
(1) a SOTA action recognition model~\cite{memvit} trained on full videos and tested on partially-observed videos in a batched fashion (i.e. we run one forward pass over the video observed so far);
(2) an adaptation of (1) for early action recognition: we re-trained the model to classify the action at each observation ratio, where we extracted features from each video segment, perform classification, and use average pooling to aggregate the predictions;
(3) our method with a transformer decoder, dynamic loss and prototype learning, as described in Section~\ref{sec:our_method}.
As reported in Table~\ref{tab:rec_ablation}, method (2) achieves better performance compared to (1) at $\rho = 12.5\%$ and slightly better at $\rho = 25\%$, but yields a big drop in AUC. With our method (3), we were able to improve the prediction accuracy at early stage up to $\rho = 62.5\%$ which leads to a significant gain on the overall AUC ($+3.33$). At $\rho = 100\%$, our method largely bridged the accuracy gap between standard action recognition and early action recognition methods, reducing it from $15.48$ to $2.14$. Figure~\ref{fig:main_fig} visualizes the accuracy plot of experiments (1) and (3).

\begin{table}[t]
    \centering
    \begin{adjustbox}{width=\linewidth, center}
	\setlength{\tabcolsep}{4pt}
    \begin{tabular}{l c c c c c c c} \toprule
    \multirow{2}{*}{\textbf{Model}} & \multirow{2}{*}{\textbf{Decoder}} &  \multirow{2}{*}{\textbf{\shortstack[c]{Inference\\ Mode}}} & \multicolumn{4}{c}{\textbf{Top-1 Acc}}  & \multirow{2}{*}{\textbf{AUC}}\\
    & & & \textbf{12.5\%} & \textbf{37.5\%} & \textbf{62.5\%} & \textbf{87.5\%} & \\
    \midrule
    MViT-B &	- & Batched & 10.21 & 23.41	& 32.93	& \cc \textbf{37.58}	& 24.94	\\
    Early-ViT & AvgPool  & Online & 18.04	& 20.41	& 22.37	& 22.98	& 18.72 \\
    Early-ViT &	T-Dec-B &  Online & \cc \textbf{25.12}	& \cc \textbf{30.36}	& \cc \textbf{33.93}	& 35.71	& \cc \textbf{28.27} \\
    \bottomrule
    \end{tabular}
    \end{adjustbox}
    \vspace{-2mm}
    \caption{\small \it Early action recognition baselines of standard recognition models on EPIC-Kitchens-55. In this table, we take a vanilla MViT-B trained for the action recognition task, and used as-is for the early-action recognition task. In the first baseline, we do not employ any decoder, and the model is used in batch mode where we perform a forward pass for each observation ratio. In the second baseline, we use an average pooling layer as simple decoder in an online fashion. In the third row we use our proposed decoder that uses a T-Dec-B model. All rows use the same encoder MViT-B. Note how the presence of a decoder is essential for improving the accuracy at early stages (+15\% at 10\% observation ratio). Note that rows 1 and 3 are also displayed in Fig.~\ref{fig:main_fig}.}
    \label{tab:rec_ablation}
    \vspace{-4mm}

\end{table}

\vspace{-2mm}
\subsection{Model Ablations}
\label{sec:ablations}
\vspace{-2mm}

\begin{table}[t]
    \centering
    \begin{adjustbox}{width=0.95\linewidth,center}
	\setlength{\tabcolsep}{3pt}
    \begin{tabular}{l c c c c c c c} 

    \toprule
    \multirow{2}{*}{\textbf{Dataset}} & \multirow{2}{*}{\textbf{Prototypes}} &  \multicolumn{5}{c}{\textbf{Top-1 Acc}}  & \multirow{2}{*}{\textbf{AUC}}\\
    & & \textbf{10\%} & \textbf{30\%} & \textbf{50\%} & \textbf{70\%} & \textbf{90\%} & \\

    \midrule
    SSsub21 & \xmark & 29.46	& 39.73	& 51.79	& 59.31	& 61.86	& 45.66	 \\
    SSsub21 & \cmark & \cc \textbf{32.08}	& \cc \textbf{40.31}	& \cc \textbf{52.36}	& \cc \textbf{60.65}	& \cc \textbf{62.24}	& \cc \textbf{46.46}	 \\
    
    \midrule
    UCF-101 &	\xmark & 91.65 &	94.45 &	95.59 &	96.01 &	96.14 &	85.66 \\
    UCF-101 & \cmark & \cc \textbf{92.15}	& \cc \textbf{95.03}	& \cc \textbf{95.75}	& \cc \textbf{96.35}	& \cc \textbf{96.54}	& \cc \textbf{85.95} \\

    \toprule
    \multirow{2}{*}{\textbf{Dataset}} & \multirow{2}{*}{\textbf{Prototypes}} &  \multicolumn{5}{c}{\textbf{Top-1 Acc}}  & \multirow{2}{*}{\textbf{AUC}}\\
    & & \textbf{12.5\%} & \textbf{37.5\%} & \textbf{50\%} & \textbf{75\%} & \textbf{87.5\%} & \\
    
    \midrule
    EK-55 & \xmark & 24.28	&	30.13	& 31.55	& 34.08	& 34.81	& 27.61	\\
    EK-55 & \cmark  & \cc \textbf{25.12}	& \cc \textbf{30.36}	& \cc \textbf{32.1}	& \cc \textbf{35.25}	& \cc \textbf{35.71}	& \cc \textbf{28.27} \\

    \bottomrule
    \end{tabular}
    \end{adjustbox}\vspace{-3mm}
    \caption{\small \it Early action recognition results on multiple datasets with and without learning prototypes.}
    \label{tab:proto_reg_abl}
    \vspace{-3mm}
\end{table}

\begin{figure}
     \centering
     \includegraphics[width=\linewidth]{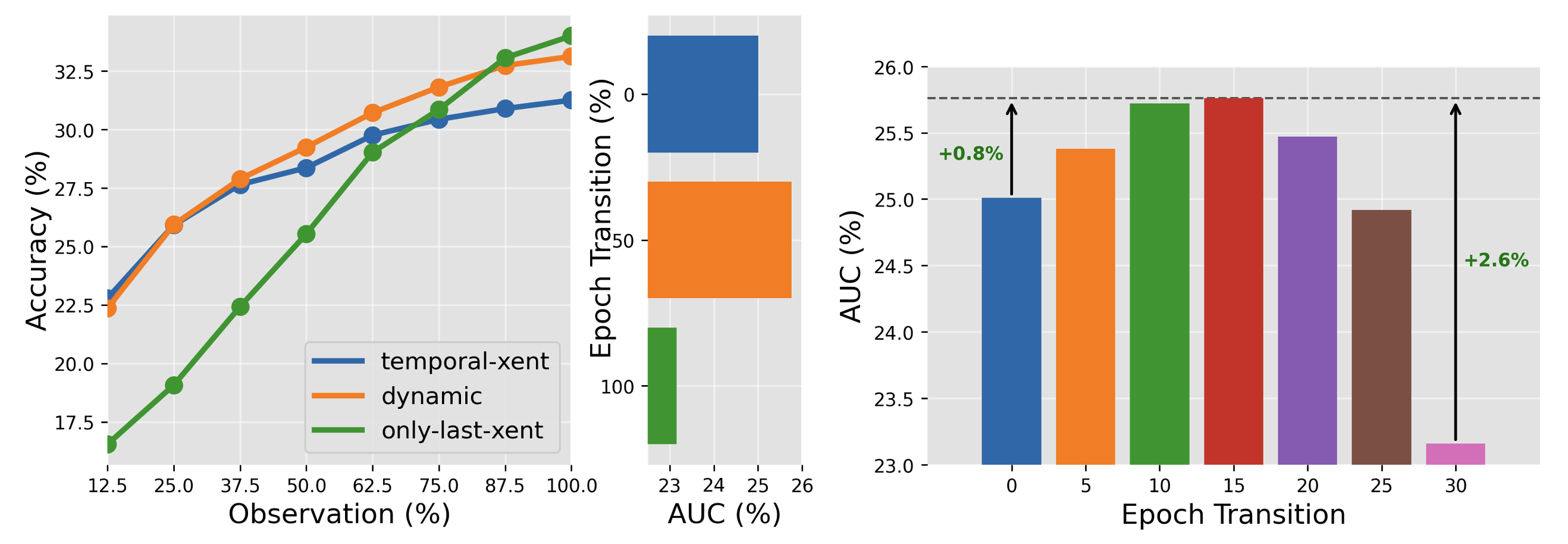}
     \vspace{-7mm}
     \caption{\small \it \textbf{Left:} Accuracy obtained with different classification losses. Standard action recognition loss ("only-last-xent") yields high accuracy when the video is fully observed, but very poor results when partially observed. In contrary, a loss that considers all the temporal tokens at the same time ("temporal-xent") exhibits the opposite behavior. Our proposed loss ("dynamic") blends the best of the two, achieving high accuracy for all the observation ratios. \textbf{Right:} Switching the loss from $\mathcal{L}_{ol}$ to $\mathcal{L}_{all}$ too late (epochs 30) results in a drop in AUC, as the model overfits towards performing well only on full-video classification.\vspace{-2mm}}
     \label{fig:loss_ablation}
     \vspace{-3mm}
\end{figure}

\begin{table}[t]
    \begin{subtable}[h]{0.45\textwidth}
    \centering
    \begin{adjustbox}{width=\linewidth,center}
	\setlength{\tabcolsep}{3pt}
    \begin{tabular}{l c c c c c c c} \toprule
    
    \multirow{2}{*}{\textbf{Model}} &  \multicolumn{5}{c}{\textbf{Top-1 Acc}} & \multirow{2}{*}{\textbf{AUC}}\\
    & \textbf{10\%} & \textbf{30\%} & \textbf{50\%} & \textbf{70\%} & \textbf{90\%} &  \\

    \midrule
    Early-ViT (no-proto) & 29.46&  39.73&  51.79&  59.31&  61.86& 45.66 \\
    Early-ViT (pred $z(t+1)$) & 30.48 &	39.99	& 52.23 & \cc \textbf{61.29} & \cc \textbf{62.24}	& 46.24 \\
    Early-ViT (pred $z(T)$) & 31.12&  39.60 &  51.98&  61.22 &  61.48 & 46.03 \\
    Early-ViT & \cc \textbf{32.08} &  \cc \textbf{40.31} &  \cc \textbf{52.36} &  60.65 &  \cc \textbf{62.24} & \cc \textbf{46.46} \\
    \bottomrule
    \end{tabular}
    \end{adjustbox}
    \caption{\small \it Model ablation on the source of regularization.} 
    \label{tab:proto_role}
    \end{subtable}
    \hfill
    \vspace{5pt}
    
    \begin{subtable}[h]{0.45\textwidth}
    \centering
    \begin{adjustbox}{width=\linewidth,center}
	\setlength{\tabcolsep}{3pt}
    \begin{tabular}{l c c c c c c c} 
    \toprule
    Early-ViT (R3D enc) & 27.74  & 38.07 &   48.15 &  54.53 &  56.76    & 42.39 \\
    Early-ViT (MViT-B enc) & \cc \textbf{32.08} &  \cc \textbf{40.31} &  \cc \textbf{52.36} &  \cc \textbf{60.65} &  \cc \textbf{62.24} & \cc \textbf{46.46} \\
    \bottomrule
    \end{tabular}
    \end{adjustbox}
    \caption{\small \it Our model trained using different encoder backbones.}
    \label{tab:encoder_abl}
    \end{subtable}
    \hfill
    \vspace{5pt}
    
    \begin{subtable}[h]{0.45\textwidth}
    \centering
    \begin{adjustbox}{width=\linewidth,center}
	\setlength{\tabcolsep}{3pt}
    \begin{tabular}{ l @{\hspace{9\tabcolsep}} c c c c c c c} \toprule
     Early-ViT ($F_c = 1$) & 27.61   &   31.95   &   37.56   &   43.75   &   46.05    &  34.74 \\
    Early-ViT ($F_c = 2$) &  32.08 &   40.31 &   52.36 &   60.65 &   62.24 &  46.46 \\
    Early-ViT ($F_c = 4$) & \cc \textbf{32.91}   &  \cc \textbf{44.01}   &   \cc \textbf{57.14}  &   \cc \textbf{66.84}  &   \cc \textbf{69.45}    &  \cc \textbf{50.80} \\
    \bottomrule
    \end{tabular}
    \end{adjustbox}
    \caption{\small \it Varying the number of frames per clip. }
    \label{tab:num_frames_enc_abl}
    \end{subtable}\vspace{-2mm}
    \caption{\small \it Ablation studies on the SSsub21 dataset. In (a) we see that the usage of prototype learning improves over a baseline without it, and also outperforms knowledge-distillation regularization techniques; in (2) the accuracy improves with stronger encoders; in (3) more frames per clip improve the accuracy.\vspace{-7mm}}
    
\end{table}

\tinyp[1mm]{Effect of prototype learning}
To demonstrate the effectiveness of learning prototypes for early action recognition, we conducted ablation study on multiple datasets with and without using prototypes. As shown in Table~\ref{tab:proto_reg_abl}, we observe \textit{consistent} accuracy improvements on \textit{all observations} ranging from around $0.5$ to $3.0$ points across datasets when using prototypes. This proves the effectiveness and generalization ability of our method.

\tinyp[1mm]{Dynamic loss}
As introduced in Sec.~\ref{sec:our_method} we found essential to design a proper dynamic loss $\mathcal{L}_{dyn}$ (see Eq.~\ref{eq:L_dyn}) to regulate at training time the learning of the prototypes while  balancing the accuracy at early stages (driven by the $\mathcal{L}_{all}$ loss in Eq.~\ref{eq:L_all}) and late predictions (driven by the $\mathcal{L}_{ol}$ loss in Eq.~\ref{eq:L_ol}).
Left figure in Figure~\ref{fig:loss_ablation} shows this behavior. When using only $\mathcal{L}_{ol}$ (only-last-xent), we can see that the model performs very poorly at all the stages, but at 100\% observation ratio, when the video has been fully observed.
This is expected, as the loss considers only the final output token which encodes the information of the entire video, similarly as what is done normally for the standard action recognition task.
In contrary, using only $\mathcal{L}_{all}$ (temporal-xent) greatly improves the accuracy at early stage, but yields sub-par results at later stages, in particular when the video is fully observed. While this is expected by design (as in Eq.~\ref{eq:L_all} we just average the losses from the individual output tokens), it is not desirable in a real-world application, as we would like the algorithm to yield the highest possible accuracy throughout.
Our dynamic loss $\mathcal{L}_{dyn}$ (dynamic) blends the best of both worlds, achieving better accuracy as $\mathcal{L}_{all}$ for all the stages, while yielding just 0.5\% less accuracy for full-video classification.
Moreover, in the right Figure of Fig.~\ref{fig:loss_ablation} we quantitatively study when it is best to switch the loss, which justifies the usage of $e^*=15$.
We also tried a soft version of $\mathcal{L}_{dyn}$, where the switching is done smoothly by using a sigmoidal weighting $\mathcal{L}_{dyn} = w(t)\mathcal{L}_{ol} + (1 - w(t))\mathcal{L}_{all}$ 
where $w(t)=\sigma(-\alpha*(t - t^*))$ and $\alpha$ is the smoothness coefficient and $t^*$ is the switching point, but obtained inferior results.

\tinyp[1mm]{Source of regularization}
In Table~\ref{tab:proto_role} we quantify the benefit in term of accuracy derived by the usage of the prototypes as regularization method, and compare it with alternative sources of regularization. As it can be seen, learning the prototypical representation of the actions improves the accuracy at every partial observation ratio, in particular at the early stages ($+2.62$ Top-1 accuracy points at 10\% observation). 
In rows "\textit{pred} $z(t+1)$" and "\textit{pred} $z(T)$", we explore the usage of knowledge-distillation as form of regularization,  where instead of learning the prototypes, for each input clip we ask the model to predict the embedding for the future next clip or for the final clip respectively as done similarly in previous works~\cite{Wang2019ProgressiveTL,Gammulle2019PredictingTF,Zhao2019SpatiotemporalFR}. We found both of these models yields inferior results compared to our usage of the prototypes.

\tinyp[2mm]{Encoder backbone model}
In Table~\ref{tab:encoder_abl} we study the effect of different encoder backbone models. We compare MViT against a standard ResNet-3D. We note that while our design can accommodate any encoder, the model does take advantage of stronger encoders, which translates into higher accuracy for all the observation ratios.

\tinyp[2mm]{Number of frames per clip}
In Table~\ref{tab:num_frames_enc_abl} we study how the number of frames per clip influences the end accuracy. We note that more frames lead to substantially higher accuracy and AUC for all observation ratios (AUC: $34.74\rightarrow46.46\rightarrow50.80$; interestingly, the accuracy does not seem to saturate at our maximum value $F_c=4$, suggesting that the performance could improve further with a larger value of $F_c$. Nevertheless, in all our previous results we used $F_c=2$ to keep the training computational cost manageable.

\begin{figure}
     \centering
     \includegraphics[width=1.0\linewidth]{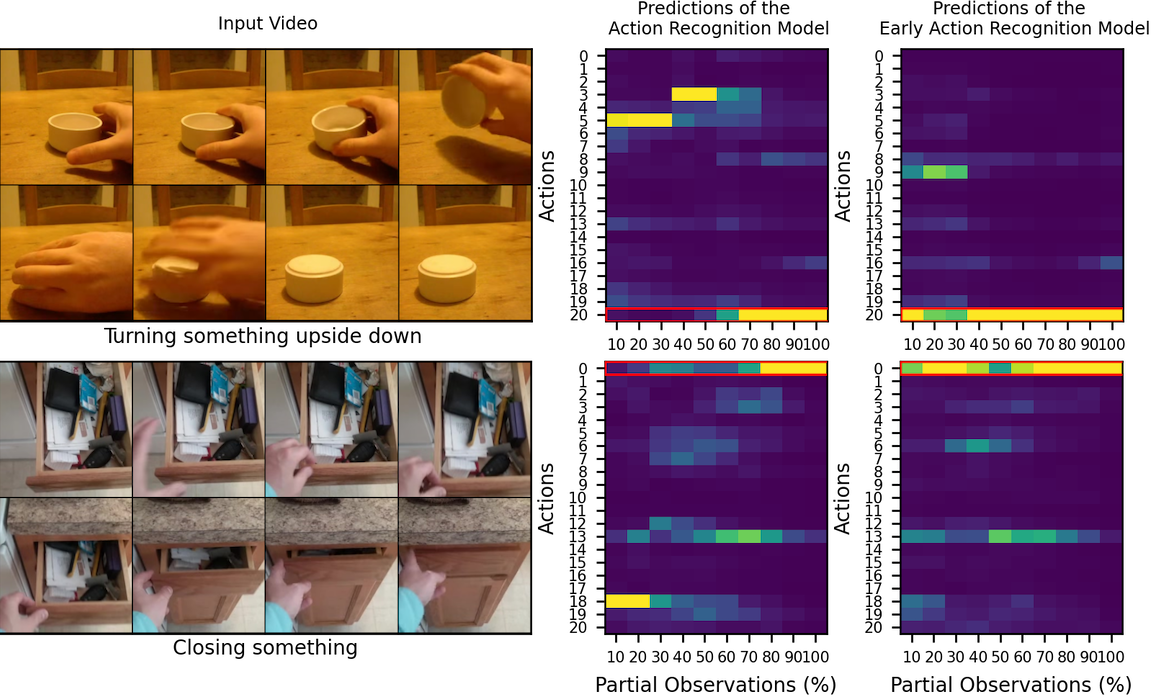}
     \vspace{-6mm}
    \caption{\small \it Visualization of classification scores at different observation ratios from standard and early action recognition models.}
    \label{fig:qual}
    \vspace{-5mm}
\end{figure}

\tinyp[2mm]{Visualization of model predictions}
In Figure~\ref{fig:qual}, we visualize the prediction scores of our early action recognition model (right) and a standard action recognition model (left) at different observation ratios. In each row, the left side shows the collage of frames from the video segments and the right side is the score visualization, where the GT class is highlighted with a red box.
We show that our model is able to correctly classify the action at early stage while the standard action recognition model needs to see more of the video before it can make the right prediction. 

\vspace{-1mm}
\section{Conclusion and future directions}
\label{sec:conclusion}
We presented Early-ViT, a novel model for early video action recognition that achieves state-of-the-art accuracy on multiple challenging real-world datasets and outperforms the prior work. 
We presented exhaustive ablation studies and qualitative visualizations to motivate our technical choices and gather insights on what the model learned.
An future extension is to use datasets with longer videos clips, where actions are complex and that could span minutes.

\clearpage
\FloatBarrier

{
    \small
    \bibliographystyle{ieeenat_fullname}
    \bibliography{main}
}

\clearpage
\title{}
\input{supplementary.tex}

\end{document}

%% file: supplementary.tex
\appendix
\begin{center}{\bf \Large Appendix}\end{center}

\vspace{12mm}
\section{Training step pseudocode}

The training step implementation, including stop-gradient operations, is presented in PyTorch-like code in Algorithm~\ref{algo:train_step}.

\vspace{6mm}
\begin{lstlisting}[language={Python}, caption={Single training step of our model.}, label={algo:train_step}, frame=tb, rulecolor=\color{gray}]
"""
encoder:  model visual encoder
proj:     encoder linear projector
decoder:  model transformer decoder
head:     classification head
xent_dyn: dynamic cross-entropy loss
cdist:    cosine similarity
f:        feature predictie module
P:        prototypes [K, D], randomly initialized
B:        batch size, T: num of segments
D:        feature dim, K: num of classes
"""
from torch import cdist
    
def train_step(x, y): # x: videos, y: labels

  # classification
  z_enc = proj(encoder(x)) # [B, T, D]
  z = decoder(z_enc) # [B, T, D]
  logits = head(z) # [B, K, T]
  preds = logits.softmax(1) # [B, K, T]
  loss_clf = xent_dyn(logits, y)
  
  # prototypes learning
  sim_P = - cdist(P, z[:, -1].detach()) # [B, K]
  loss_proto = xent(sim_P, y)
  
  # prototypes prediction
  z_feat = f(z) # [B, T, D]
  sim_feat = - cdist(z_feat, P.detach()) # [B, K, T]
  loss_feat = xent(sim_feat, y)
  
  # total loss
  loss = loss_clf + loss_proto + loss_feat
  return preds, loss
\end{lstlisting}
\vspace{4mm}

\section{Additional implementation details}
\vspace{2mm}
Table~\ref{tab:models_info} presents detailed information regarding the encoder and decoder input/output tensor shapes of our model.

\begin{table}[!ht]
    \centering
    \begin{adjustbox}{width=\linewidth,center}
	\setlength{\tabcolsep}{4pt}
    \begin{tabular}{l c c c c c c} \toprule
    \textbf{Module} & \textbf{Instance} & \textbf{Size} & \textbf{Depth} & \textbf{D} & \begin{tabular}{@{}c@{}}\textbf{Input} \\ \textbf{Shape}\end{tabular} & \begin{tabular}{@{}c@{}}\textbf{Output} \\ \textbf{Shape}\end{tabular}\\
    \midrule
    Encoder & MViT~\cite{Fan2021MultiscaleVT} & Base & 16 & 768 & $[C, F_{seg} , H, W]$ & $D$\\
    Decoder & T-Dec~\cite{NIPS2017_3f5ee243} & Base & 6 & 768 & $[N_{seg}, D]$ & $[N_{seg}, D]$\\

    \bottomrule
    \end{tabular}
    \end{adjustbox}
    \caption{Encoder and decoder details.}
    \label{tab:models_info}
\end{table}

\vspace{2mm}
\section{Comparison with TemPr}
In this section, we conduct an in-depth comparison with TemPr~\cite{Stergiou_2023_CVPR} which is the previous SOTA on early action recognition. To provide an apple-to-apple comparison, we trained Early-ViT with the same backbone as the ones that produce the best result for TemPr~\cite{Stergiou_2023_CVPR} on three datasets SSv2, SSsub21 and UCF-101.
Specifically, Table~\ref{tab:tempr_backbone_ablation} presents the detailed results of TemPr~\cite{Stergiou_2023_CVPR} and our method when using different backbones. We found that both methods have better compatibility with a certain encoder type. However, our method is able to achieve better performance with the most compatible backbone given far less amount of frames.

\begin{table*}[!htp]
    \centering
    \begin{adjustbox}{width=\linewidth,center}
	\setlength{\tabcolsep}{4pt}
    \begin{tabular}{l l c c c c c c c c c c c c c c c} \toprule
    \multirow{2}{*}{\textbf{Dataset}} & \multirow{2}{*}{\textbf{Model}} & \multirow{2}{*}{\textbf{Encoder}} & \multirow{2}{*}{\textbf{\shortstack[c]{Inference\\ Mode}}} & \multicolumn{10}{c}{\textbf{Top-1 Acc}}  & \multirow{2}{*}{\textbf{AUC}} & \multirow{2}{*}{\textbf{\shortstack[c]{\# Frames at\\ $\rho=100\%$}}} & \multirow{2}{*}{\textbf{\shortstack[c]{GFLOPs at\\ $\rho=100\%$}}}\\
    & & & & \textbf{10\%} & \textbf{20\%} & \textbf{30\%} & \textbf{40\%} & \textbf{50\%} & \textbf{60\%} & \textbf{70\%} & \textbf{80\%} & \textbf{90\%} & \textbf{100\%} & \\
    \midrule
    \multirow{3}{*}{\textbf{SSv2}} & TemPr~\cite{Stergiou_2023_CVPR} & V-Swin-B & Batched & 20.5 &	-	& 28.6	& -	& 41.2	& -	& 47.1	& - &	-	& -	& - & $4\times 16\times 10$ & - \\
    & Early-ViT (ours)&	V-Swin-B & Online & \cc \textbf{23.34}	& \underline{27.0}	& \underline{31.93}	& \underline{37.84}	& \underline{44.47}	& \underline{50.67}	& \underline{55.35}	& \underline{58.81}	& \underline{60.75}	& \underline{61.33}	& \underline{40.92}	 & $2\times 10$ & - \\
    & \cc Early-ViT (ours)&	\cc MViT-B & \cc Online & \underline{22.73}	& \cc \textbf{27.81}	& \cc \textbf{33.62}	& \cc \textbf{40.52}	& \cc \textbf{47.95}	& \cc \textbf{53.94}	& \cc \textbf{58.54}	& \cc \textbf{61.49}	& \cc \textbf{63.03}	& \cc \textbf{63.56}	& \cc \textbf{43.00}	& $2\times 10$ & - \\
    \midrule
    \multirow{3}{*}{\textbf{SSsub21}} & TemPr~\cite{Stergiou_2023_CVPR} & MoViNet-A4 & Batched & \underline{28.4} & \underline{34.8} & \underline{37.9} & - & 41.3 & - & 45.8 & - & 48.6 & - & - & $4\times 16\times 10$ & - \\
    & Early-ViT (ours) & MoViNet-A4 & Online & 23.9 & 28.8 & 33.3 & \underline{38.9} & \underline{45.0} & \underline{48.1} & \underline{51.9} & \underline{54.0} & \underline{54.5} & \underline{54.7} & \underline{39.38}	& $2\times 10$ & -  \\
    & \cc Early-ViT (ours)&	\cc MViT-B  & \cc Online & \cc \textbf{32.08}	& \cc \textbf{35.52}	& \cc \textbf{40.31}	& \cc \textbf{47.39}	& \cc \textbf{52.36}	& \cc \textbf{57.14}	& \cc \textbf{60.65}	& \cc \textbf{61.86}	& \cc \textbf{62.24}	& \cc \textbf{62.12}	& \cc \textbf{46.46}	& $2\times 10$ & - \\
    \midrule
    \multirow{4}{*}{\textbf{UCF-101}} & TemPr~\cite{Stergiou_2023_CVPR} &	MoViNet-A4  &	Batched & \underline{88.6} &	\underline{93.5} &	\underline{94.9} &	\underline{94.9} &	\underline{95.4} &	\underline{95.2} &	\underline{95.3} &	\cc \underline{\textbf{96.6}} &	\underline{96.2} &	- &	- & $4\times 16\times 10$ & - \\
    & Early-ViT (ours) & MoViNet-A4 & Online & 88.11 & 90.83 & 92.2 & 92.71 & 93.42 & 93.66 & 93.92 & 93.68 & 94.08 & 93.84 & 83.55 & $5\times 10$ & - \\
    & TemPr$^*$~\cite{Stergiou_2023_CVPR} &	MViT-B & Batched &	85.7 &	- &	87.4 &	- &	87.5 &	- &	86.9 &	- &	86.5 &	- &	- & $4\times 16\times 10$ & 291 \\
    & \cc Early-ViT (ours) &	\cc MViT-B & \cc Online & \cc \textbf{92.15}	& \cc \textbf{93.97}	& \cc \textbf{95.03}	& \cc \textbf{95.3}	& \cc \textbf{95.75}	& \cc \textbf{95.88}	& \cc \textbf{96.35} &  96.33	& \cc \textbf{96.54}	& \cc \textbf{96.64}	& \cc \textbf{85.95} & $2\times 10$ & 15.6 \\
    \bottomrule
    \multicolumn{17}{l}{\small * indicates our implementation of TemPr~\cite{Stergiou_2023_CVPR} using their official code. We trained MViT-B with the configuration reported in their paper which they trained multiple backbones with.}
    \end{tabular}
    \end{adjustbox}
    \caption{\it Comparison between Early-ViT and TemPr~\cite{Stergiou_2023_CVPR} on multiple datasets with different backbones. Overall our method is able to achieve better performance with the most compatible backbone given far less amount of frames and computation. Our method also generalizes better to other backbones than TemPr~\cite{Stergiou_2023_CVPR} with a smaller drop in accuracy when swapping backbone with TemPr~\cite{Stergiou_2023_CVPR} on UCF-101.}
    \label{tab:tempr_backbone_ablation}
\end{table*}

\begin{figure*}[!ht]
     \centering
     \includegraphics[width=\linewidth]{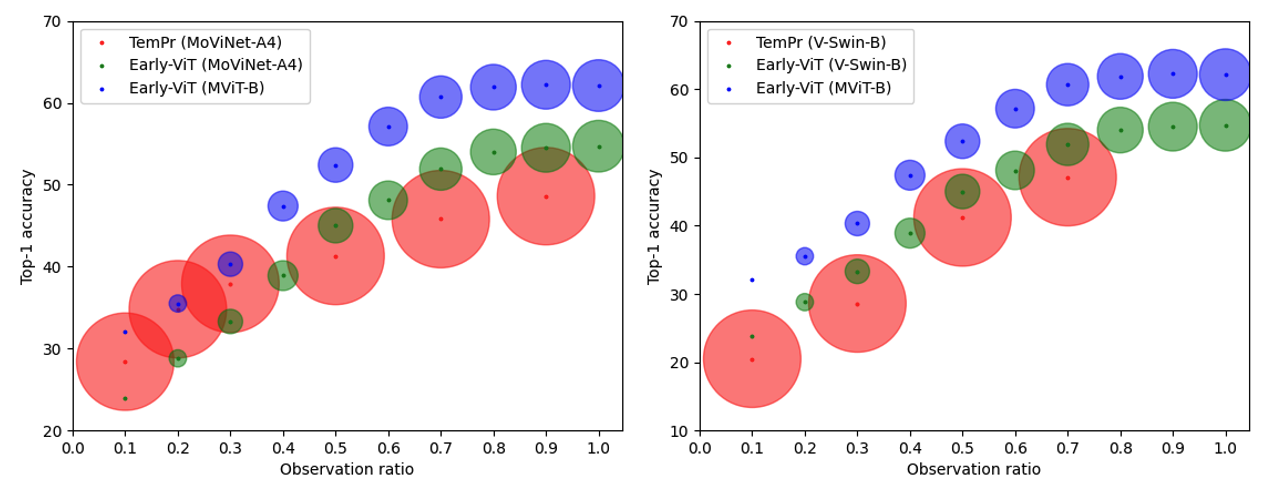}
     \caption{\it Top-1 accuracy vs the number of frames used for each observation ratio on SSsub21 dataset (\textbf{Left}) and SSv2 dataset (\textbf{Right}). The size of each circle correlates to the number of frames each model uses for training and inference.}
     \label{fig:num_frame_acc_plot}
\end{figure*}

On SSv2, TemPr~\cite{Stergiou_2023_CVPR} reports the best results with V-Swin-B. After retraining our method with V-Swin-B, we found our method still outperforms TemPr~\cite{Stergiou_2023_CVPR} with around 3\% to 8\% absolute improvement in top-1 accuracy. Our most compatible backbone, MViT-B, further brings another 1\% to 3\% improvement. On SSsub21, TemPr~\cite{Stergiou_2023_CVPR} shows the best performance with MoViNet-A4 as backbone. Our method is on par with TemPr~\cite{Stergiou_2023_CVPR} with the same backbone MoViNet-A4. We observed around 4\% to 6\% accuracy improvement starting at 50\% observation ratio while a similar regression before 50\%. However, it's worth noting that Early-ViT uses significantly less amount of frames than TemPr~\cite{Stergiou_2023_CVPR} during both training and inference, especially when the observation ratio is low. For example, when $\rho = 10\%$, TemPr~\cite{Stergiou_2023_CVPR} samples $4 \times 16 = 64$ frames with repetition while we use only 2 unique frames. Figure~\ref{fig:num_frame_acc_plot} provides a more complete picture of the accuracy each method achieves w.r.t. the number of frames it uses during training and inference. Our method uses much less frames throughout different observation ratios while still achieving on par or even better performance. Furthermore, as pointed out in the main paper, our method trains only one model end-to-end while TemPr~\cite{Stergiou_2023_CVPR} trains one model per observation model. Therefore, our method is much more efficient in terms of both training and inference.

On UCF-101, TemPr~\cite{Stergiou_2023_CVPR} reports the best results with MoViNet-A4. After retraining our method with MoViNet-A4 on UCF-101, we found a regression of around 1\% to 3\% in top-1 accuracy. To understand TemPr's~\cite{Stergiou_2023_CVPR} compatibility with our backbone, we also retrained TemPr~\cite{Stergiou_2023_CVPR} with MViT-B. We found a larger regression of around 8\% to 10\% in top-1 accuracy. This indicates that our method generalizes better to different backbones. When combined with the most compatible backbone, our method can achieve better performance than the best configuration of TemPr~\cite{Stergiou_2023_CVPR}.

\vspace{2mm}
\section{Qualitative Results}

\begin{figure*}[!ht]
    \centering
    
    \begin{subfigure}[h]{\textwidth}
     \includegraphics[width=0.98\linewidth]{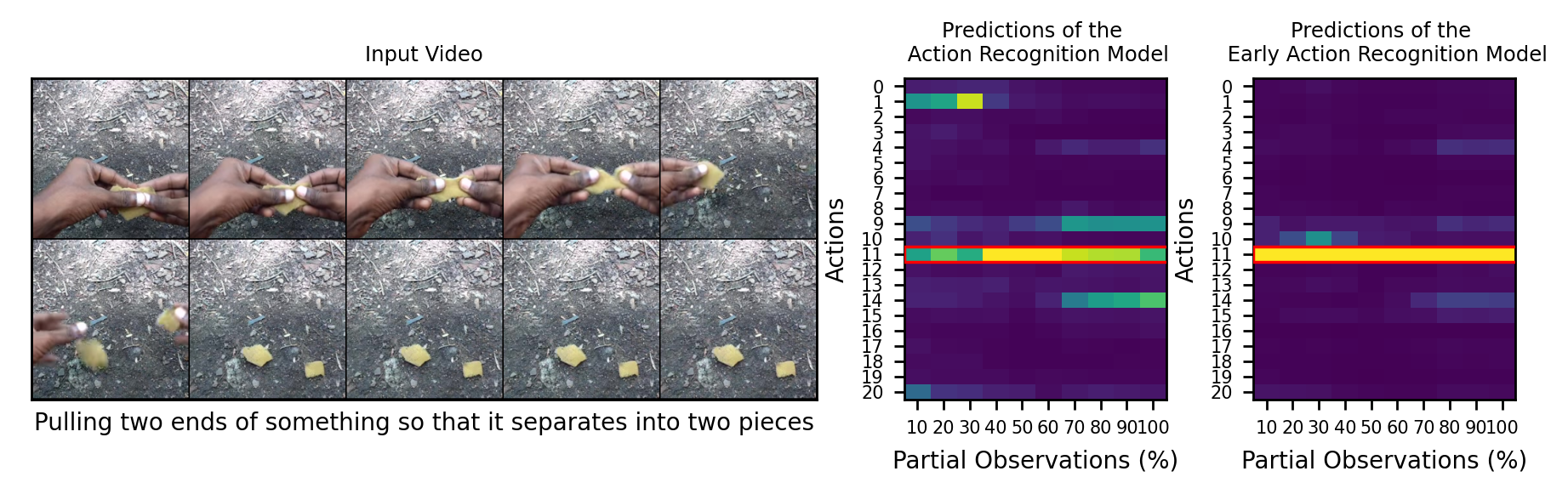}
     \label{fig:qual_40}
     \end{subfigure}
     \vspace{-3mm}
     
    \begin{subfigure}[b]{\textwidth}
     \includegraphics[width=0.98\linewidth]{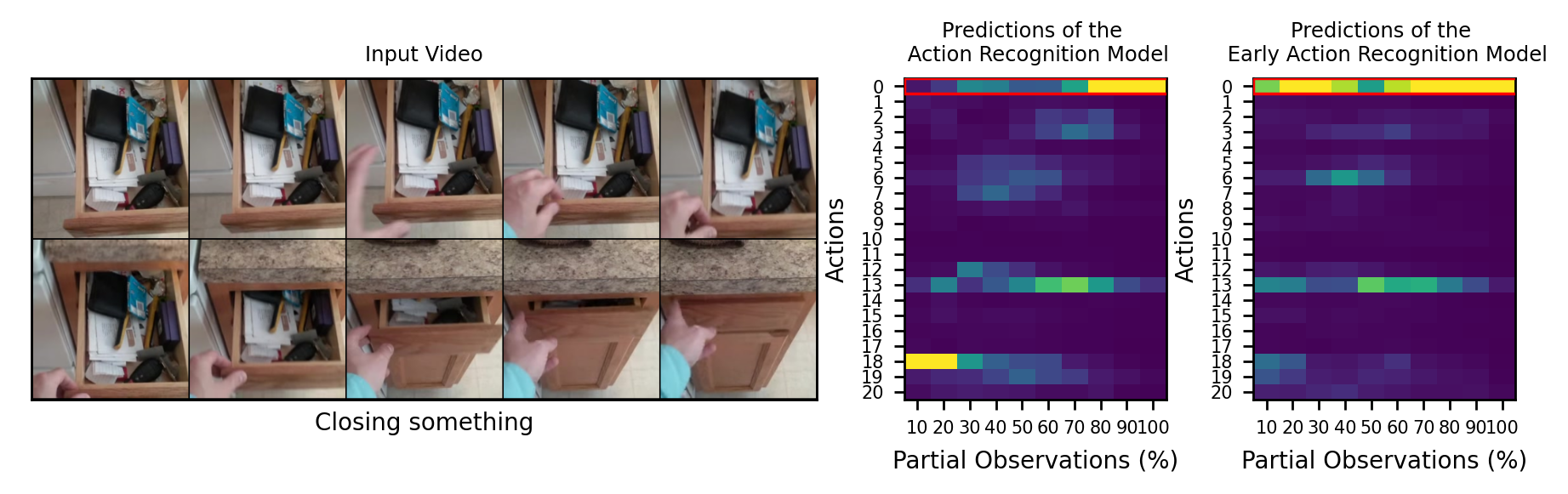}
     \label{fig:qual_502}
    \end{subfigure}
    \vspace{-7mm}
    
    \begin{subfigure}[b]{\textwidth}
     \includegraphics[width=0.98\linewidth]{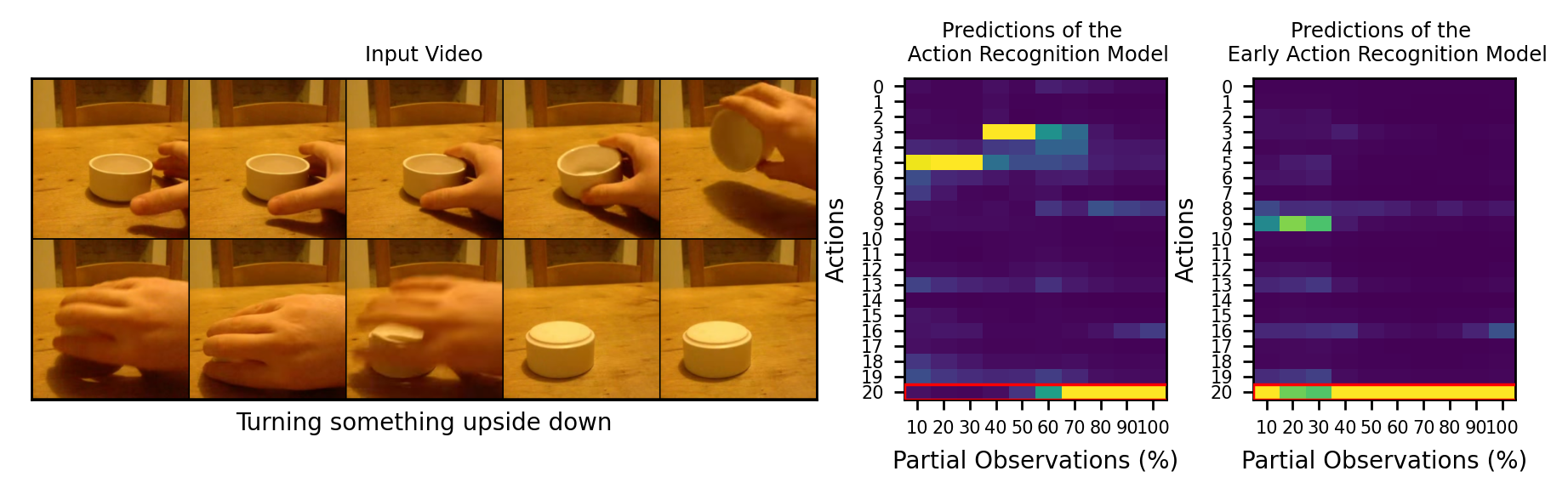}
     \label{fig:qual_154}
    \end{subfigure}
    \vspace{-7mm}
    
    \begin{subfigure}[b]{\textwidth}
     \includegraphics[width=0.98\textwidth]{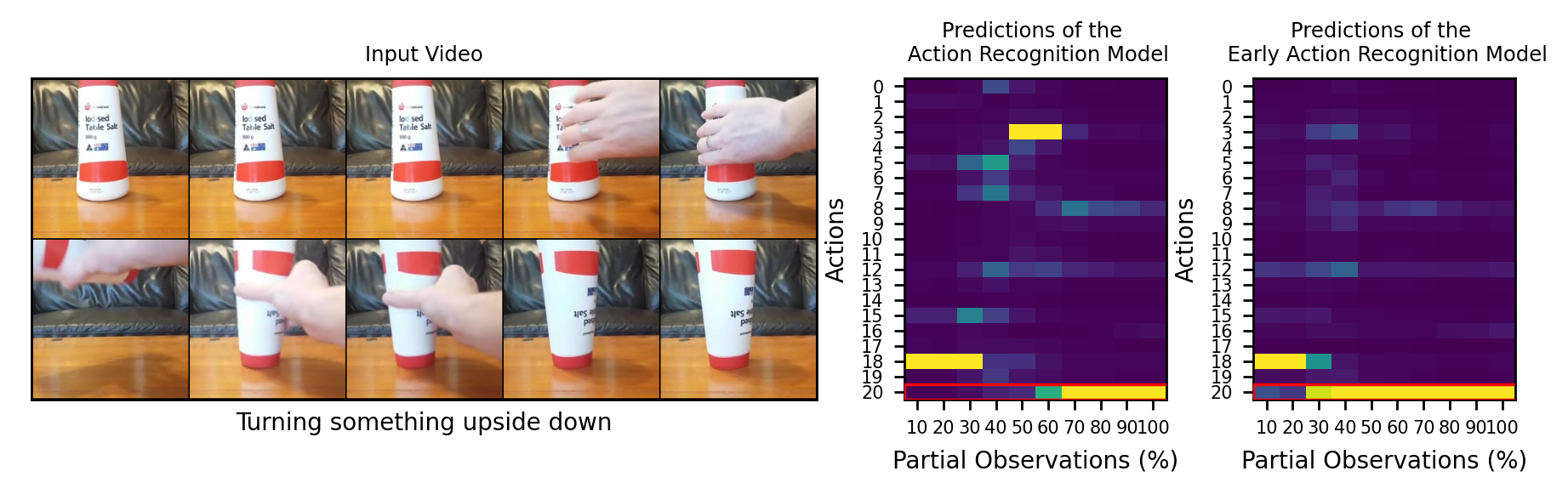}
     \label{fig:qual_1002}
    \end{subfigure}
    \vspace{-9mm}
    
    \caption{Visualization of classification scores at different observation ratios from standard and early action recognition models. Please refer to Table~\ref{tab:qual:label_descriptions} for mapping class indexes to their textual descriptions.}
    \label{fig:qual_suppl}
\end{figure*}

\begin{figure*}[ht!]
    \begin{minipage}[t]{0.47\linewidth}
        \vspace{0pt}
        \includegraphics[width=\linewidth]{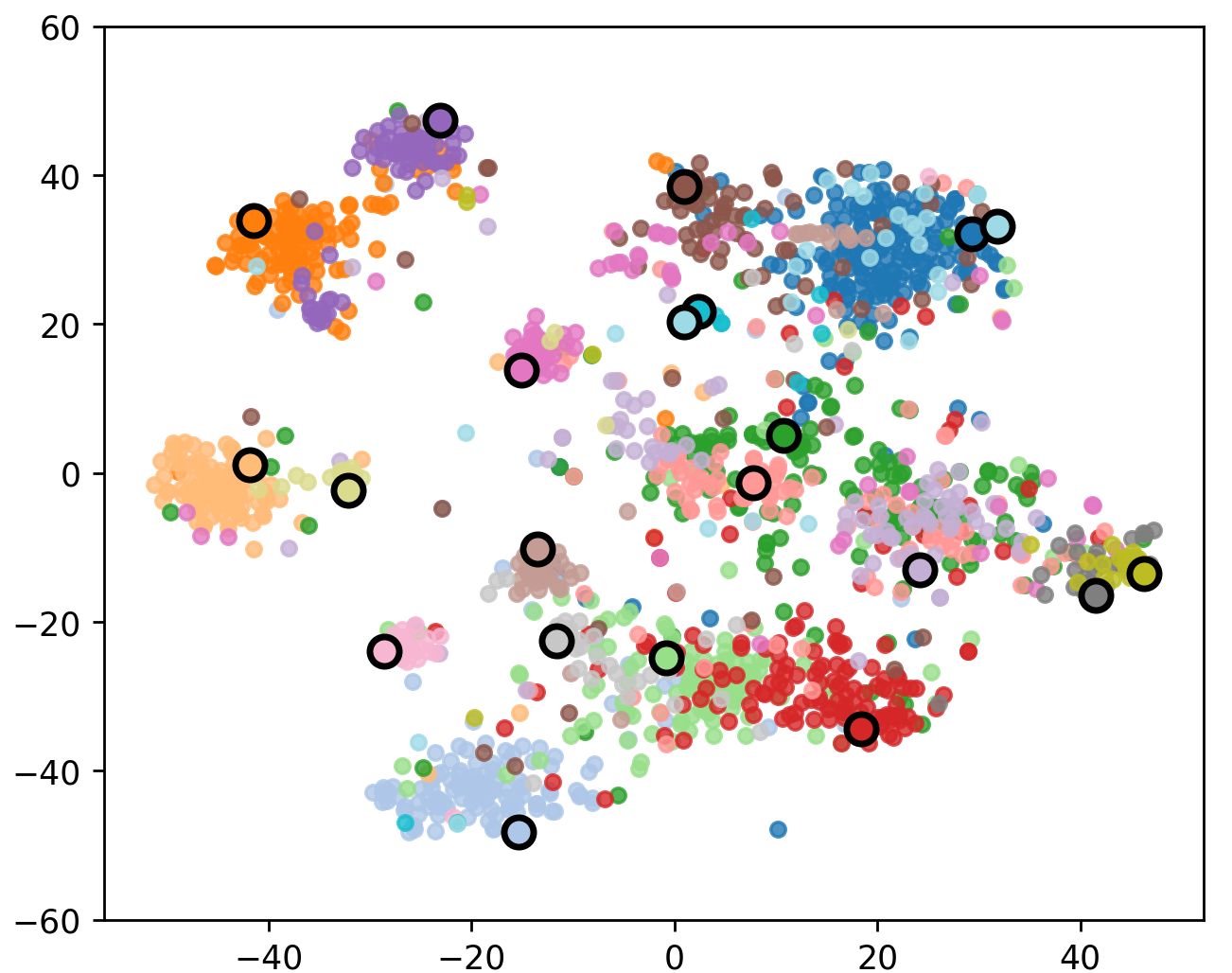}
        \caption{t-SNE visualization of the learned prototypes and test examples for the dataset SSsub21.}
        \label{fig:tsne_sssub21_proto}
    \end{minipage}
    \hspace{0.3cm}
    \begin{minipage}[t]{0.48\linewidth}
        \vspace{8pt}
        \resizebox{\linewidth}{!}{%
        \setlength{\tabcolsep}{2pt}
        \begin{tabular}{c c l}
        \toprule
        \textbf{Idx} & \textbf{Color} & \textbf{Class Description} \\
        \midrule
        0 & \textcolor[RGB]{93,144,193}{\fdot} & Closing something \\
        1 & \textcolor[RGB]{194,209,234}{\fdot} & Holding something \\ 
        2 & \textcolor[RGB]{241,157,82}{\fdot} & Opening something \\
        3 & \textcolor[RGB]{246,202,154}{\fdot} & Picking something up \\
        4 & \textcolor[RGB]{110,177,97}{\fdot} & Poking a stack of something so the stack collapses \\
        5 & \textcolor[RGB]{184,227,167}{\fdot} & Poking a stack of something without the stack collapsing \\
        6 & \textcolor[RGB]{206,92,88}{\fdot} & Pretending to open something without actually opening it \\
        7 & \textcolor[RGB]{243,176,173}{\fdot} & Pretending to sprinkle air onto something \\
        8 & \textcolor[RGB]{163,134,198}{\fdot} & Pretending to turn something upside down \\
        9 & \textcolor[RGB]{205,192,219}{\fdot} & Pulling something from left to right \\
        10 & \textcolor[RGB]{156,122,113}{\fdot} & Pulling two ends of something so that it gets stretched \\
        11 & \textcolor[RGB]{202,177,170}{\fdot} & Pulling two ends of something so that it separates into two pieces \\
        12 & \textcolor[RGB]{220,150,203}{\fdot} & Pushing something from left to right \\
        13 & \textcolor[RGB]{140,198,218}{\fdot} & Putting something into something \\
        14 & \textcolor[RGB]{152,152,152}{\fdot} & Putting something on the edge of something so it is not supported and falls down \\
        15 & \textcolor[RGB]{210,210,210}{\fdot} & Putting something similar to other things that are already on the table \\
        16 & \textcolor[RGB]{201,202,98}{\fdot} & Putting something upright on the table \\
        17 & \textcolor[RGB]{226,226,171}{\fdot} & Showing a photo of something to the camera \\
        18 & \textcolor[RGB]{110,200,214}{\fdot} & Showing something behind something \\
        19 & \textcolor[RGB]{187,224,233}{\fdot} & Stuffing something into something \\
        20 & \textcolor[RGB]{187,224,233}{\fdot} & Turning something upside down \\
        \bottomrule
        \end{tabular}
        }
        \captionof{table}{Labels textual descriptions for the SSsub21 test set.}
        \label{tab:qual:label_descriptions}
    \end{minipage}
\end{figure*}

\subsection{Visualization of model predictions.}
In Figure~\ref{fig:qual_suppl}, we visualize the prediction scores of our early action recognition model and the relative action recognition model (MViT) at different observation ratios. In each row, the left side shows the collage of frames from all video segments and the right side is the score visualization, where the ground truth class is highlighted with red box. All four examples demonstrate that our model is able to correctly classify the action at the early stage of the video while the standard action recognition model gets confused with other classes. Looking at the scores at the early stage, we can also see how the prediction from early action recognition model matches with the ambiguity of a partially observed video. For example, in the last row where the ground truth class is "Turning something upside down", it's not very clear what the action would be with only the first three segments observed. Our model produces moderate scores for the class "Pulling something from left to right" which resembles the visual cues in the first three segments as the hand is approaching from the right and grabbing the cup. The score of this ambiguous class immediately die down after the fourth segment is observed, where the cup is being tilted. As a comparison, the standard action recognition model fires high scores for a completely unrelated class "Poking a stack of something without the stack collapsing" instead. Please refer to Table~\ref{tab:qual:label_descriptions} for mapping class indexes to their textual descriptions.

\subsection{Prototype visualization.}
In Figure~\ref{fig:tsne_sssub21_proto} we show the t-SNE visualization for the learned prototypes and test examples for the SSsub21 dataset. In particular, different colors represent different action classes (reported in Table~\ref{tab:qual:label_descriptions}) and the prototypical representation of each class 
 is reported with edged circles.